\documentclass[letterpaper, 10 pt, conference]{ieeeconf}  

\IEEEoverridecommandlockouts                              

\usepackage{graphicx}
\usepackage{subfigure}
\usepackage{xcolor}
\usepackage{comment}
\usepackage[normalem]{ulem}

\newcommand{\specialcell}[2][c]{%
  \begin{tabular}[#1]{@{}c@{}}#2\end{tabular}}

\newcommand{\as}[1]{\textcolor{teal}{(AS: #1)}}

\newcommand{\ess}[1]{{\color[rgb]{0,.2,.5}{(ESS: #1)}}}

\newif\ifshowcomments
\showcommentstrue 

\usepackage{dcolumn}

\title{\LARGE \bf
Understanding Acoustic Patterns of Human Teachers\\ Demonstrating Manipulation Tasks to Robots
}

\author{Akanksha Saran$^{*}$$^{1}$, Kush Desai$^{*}$$^{2}$, Mai Lee Chang$^{2}$, Rudolf Lioutikov$^{3}$, Andrea Thomaz$^{2}$ and Scott Niekum$^{4}$
\thanks{$^{1}$Akanksha Saran is with Microsoft Research, New York, NY 10012, USA. Work done at the University of Texas at Austin.
        {\tt\small akanksha.saran@microsoft.com}}%
\thanks{$^{2}$Kush Desai, Mai Lee Chang and Andrea Thomaz are with the Department of Electrical and Computer Engineering, University of Texas at Austin, Austin, TX 78712, USA.
        {\tt\small kushkdesai@utexas.edu},
        {\tt\small mlchang@utexas.edu},
        {\tt\small athomaz@ece.utexas.edu}}%
\thanks{$^{3}$Rudolf Lioutikov is with the Karlsruhe Institute of Technology, Karlsruhe, BW 76131, Germany. 
        {\tt\small lioutikov@kit.edu }}
\thanks{$^{4}$Scott Niekum is with the Department of Computer Science, University of Texas at Austin, Austin, TX 78712, USA.
        {\tt\small sniekum@cs.utexas.edu}}
\thanks{$^{*}$ Equal contribution.}
}

\begin{document}

\maketitle
\thispagestyle{empty}
\pagestyle{empty}

\begin{abstract}

Humans use audio signals in the form of spoken language or verbal reactions effectively when teaching new skills or tasks to other humans. While demonstrations allow humans to teach robots in a natural way,
learning from trajectories alone does not leverage other available modalities including audio from human teachers.
To effectively utilize audio cues accompanying human demonstrations, first it is important to understand what kind of information is present and conveyed by such cues. 
This work characterizes audio from human teachers demonstrating multi-step manipulation tasks to a situated Sawyer robot using three feature types: (1) duration of speech used, (2) expressiveness in speech or prosody, and (3) semantic content of speech. We analyze these features along four dimensions and find that teachers convey similar semantic concepts via spoken words for different conditions of (1) demonstration types, (2) audio usage instructions, (3) subtasks, and (4) errors during demonstrations. However, differentiating properties of speech in terms of duration and expressiveness are present along the four dimensions, highlighting that human audio carries rich information, potentially beneficial for technological advancement of robot learning from demonstration methods.

\end{abstract}
\section{Introduction}

Human speech or audio is a natural, low-effort and rich channel of communication \cite{oviatt2004we}. Typically, robot learning from demonstration (LfD) algorithms ignore information carried by a human teacher's audio cues
, and only work with the state of the environment and 
the human teacher's actions \cite{argall2009survey, osa2018algorithmic}. Some prior works have leveraged gaze patterns of human demonstrators \cite{saran2018human,saran2020understanding,zhang2020human,saran2020efficiently}. However, to the best of our knowledge, human audio has been primarily unexplored in the context of learning from demonstration. 
A primary reason why incorporating this additional information has been challenging is the lack of understanding about how complex human audio signals are used and what they convey during demonstrations. 
Recent advancements in sensor technologies and speech processing algorithms~\cite{yu2016automatic} make it possible to extract informative features from human audio.
\textit{Raw} human audio carries more information than is present in a transcribed narration of textual words produced by an automatic speech recognition (ASR) system \cite{yu2016automatic}. Audio can contain information via spoken language, as well as out-of-vocabulary words not part of a natural-language learning corpus, disfluencies (restarts, repetitions, and self-corrections), filled pauses (`um', `uh'), hyperarticulations such as careful enunciation, slow speaking rate, increased pitch and loudness \cite{clark2002using}.
This work is a first step in analyzing 
how humans use unrestricted audio cues during demonstrations for multi-step manipulation tasks
---characterizing duration, prosodic features, and spoken words.

\begin{figure}
\centering
\subfigure[Kinesthetic Demonstration]{
\includegraphics[width=0.225\textwidth]{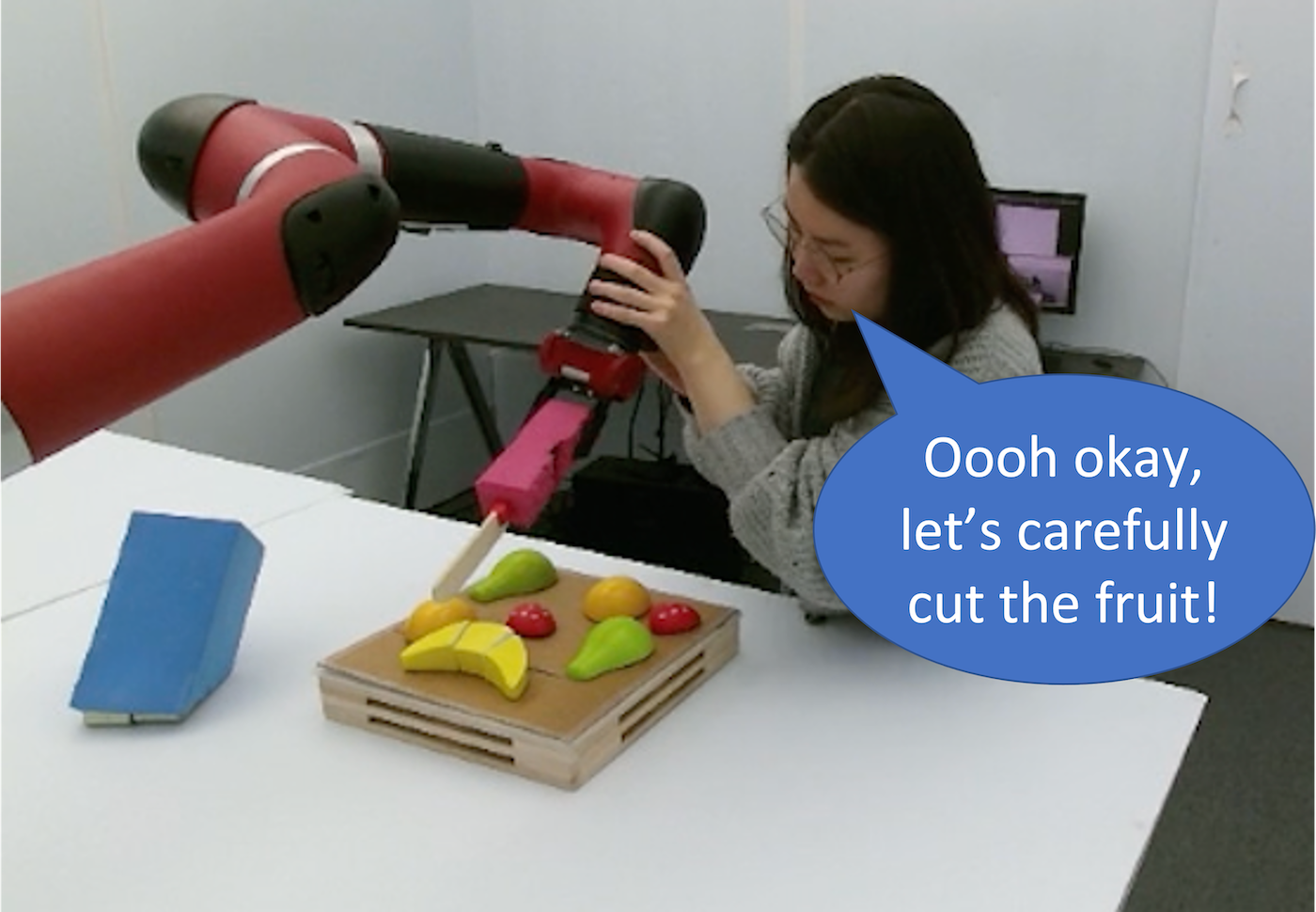}
}
\subfigure[Video Demonstration]{
\includegraphics[width=0.225\textwidth]{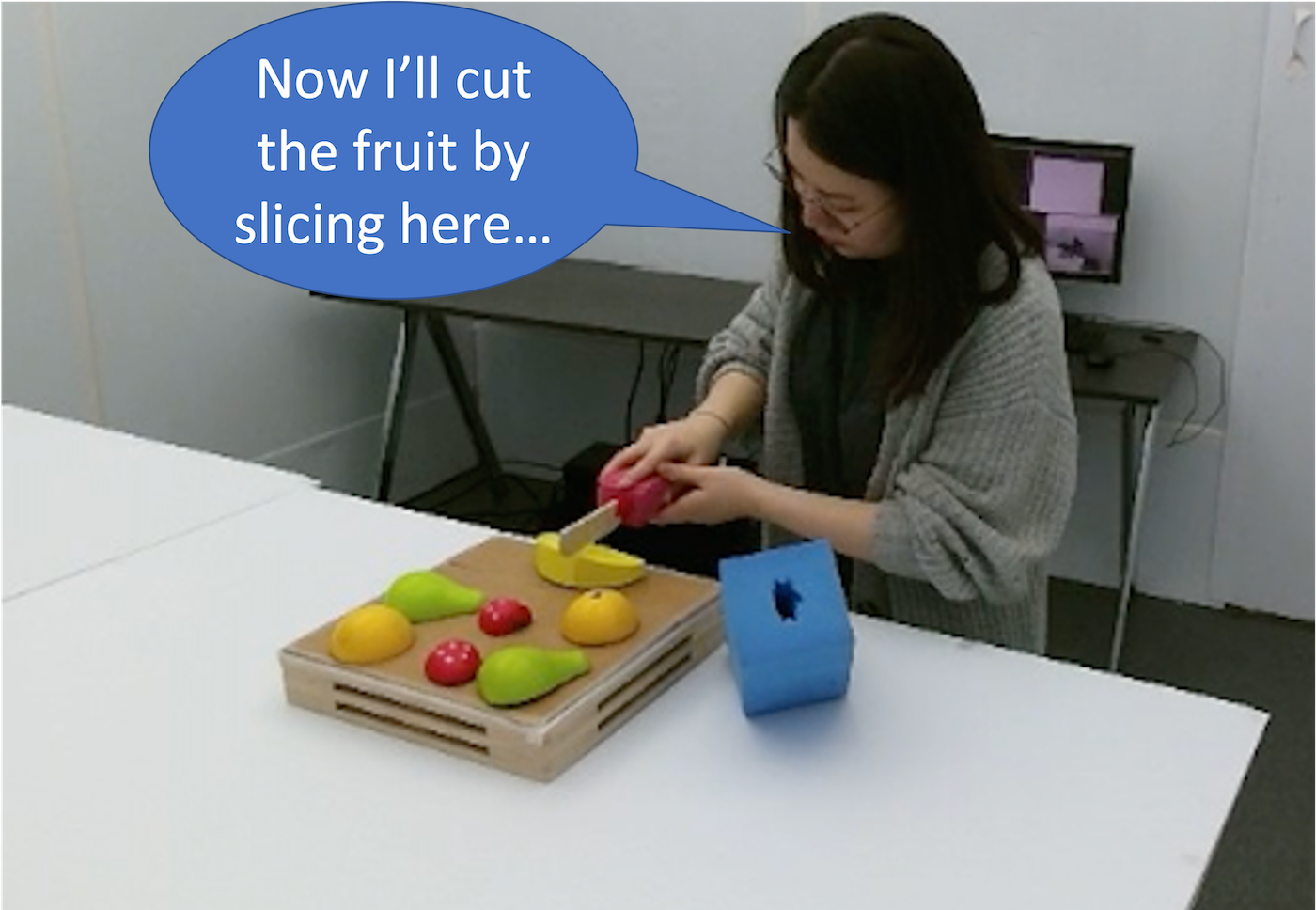}
}
\caption{
(a) Kinesthetic demonstrations and (b) Video demonstrations along with unrestricted audio signals of human teachers are recorded for two manipulation tasks. 
}
\vspace{-5mm}
\label{fig:sawyer}
\end{figure}

We work in a limited-data regime with an in-house dataset of demonstrations and accompanying human audio, collected via a human subjects study in a laboratory setting. With the help of human annotators, we characterize human speech via: (1) duration of speech in a demonstration, (2) annotated and computational prosodic features, (3) semantic content of words spoken during speech.
We analyze these features across different dimensions: (a) type of demonstration used (kinesthetic teaching using the robot's arm versus the human performing the task themselves), (b) the instruction given to a teacher about usage of speech during demonstrations (explicit narration instructions versus implicit indication to use speech), (c) presence or absence of relevant subtasks being executed, (d) presence or absence of errors during demonstrations. We find that users convey similar semantic concepts through spoken words across all four dimensions. However, we observe that teachers are more expressive but talk less densely during kinesthetic teaching compared to demonstrating the task themselves. Moreover, teachers are similarly expressive across audio usage instructions for a majority of acoustic features, but overall talk more when explicitly asked to do so.  
We also find that with a majority of acoustic features, teachers use speech more densely and more expressively during relevant subtask executions and in the absence of errors. However, some interesting speech features exist even in the absence of subtasks and in the presence of errors.

Finally, with a proof-of-concept experiment, we show that human acoustic features are useful to detect presence of relevant subtasks and errors during demonstrations. Taken together, our findings highlight that human speech carries rich information about demonstrations, which can be beneficial for technological advancement of robot learning algorithms in the future.

\begin{figure}
\centering
\subfigure[Box Opening]{
\includegraphics[width=0.08\textwidth]{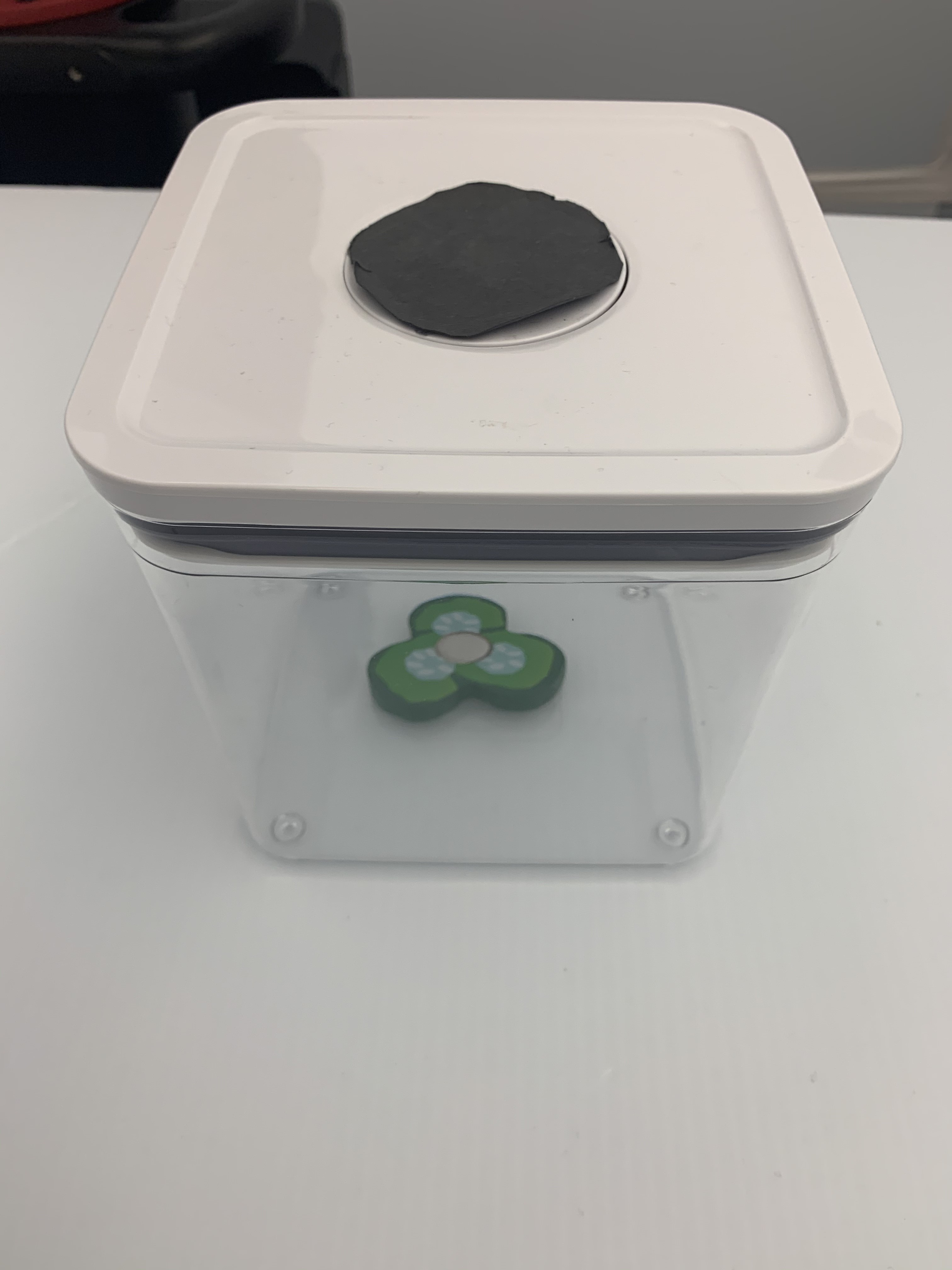}
\includegraphics[width=0.08\textwidth]{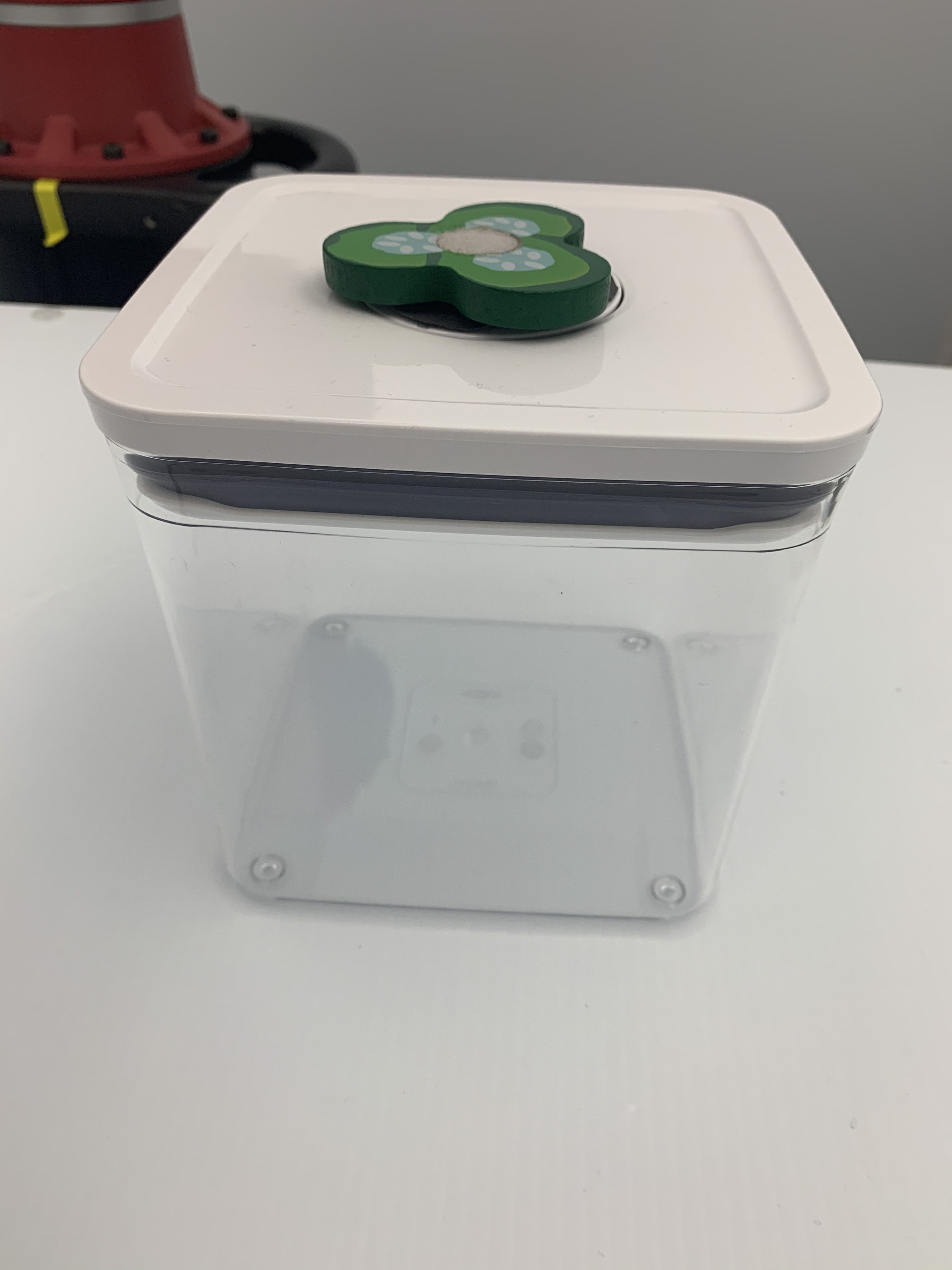}
}
\subfigure[Fruit Cutting]{
\includegraphics[width=0.13\textwidth]{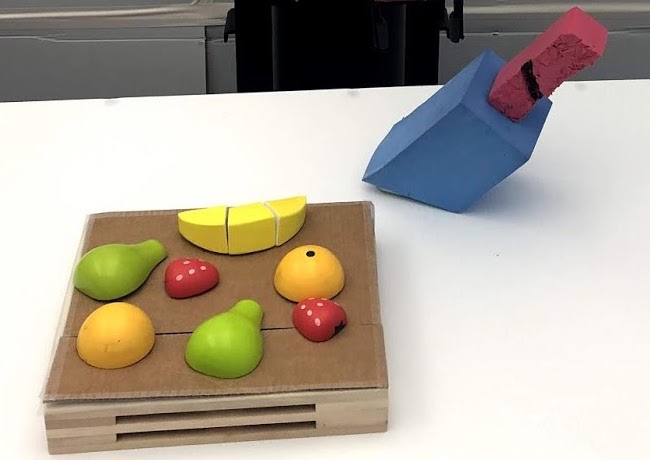}
\includegraphics[width=0.13\textwidth]{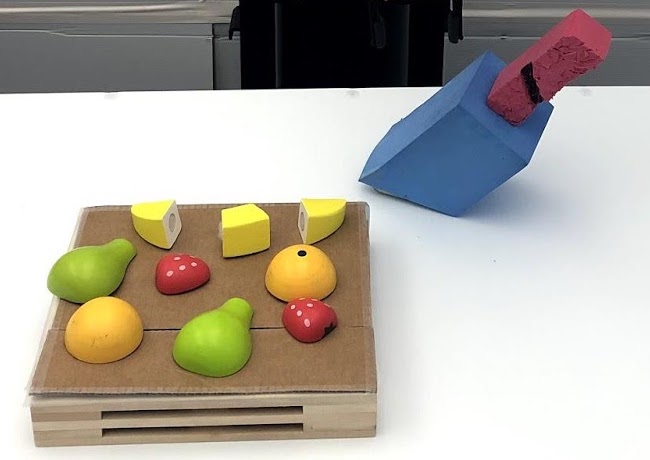}
}
\caption{Start and end conditions of two manipulation tasks for which demonstrations are provided to a Sawyer robot.}
\label{fig:sawyer-tasks}
\vspace{-5mm}
\end{figure}

\section{Related Work}

\subsection{Human Audio-assisted Human-Robot Interaction}

Scassellati et al. \cite{scassellati2009affective} showed that understanding emotion in the context of a conversation is an essential skill for keeping humans engaged when interacting with robots. They developed a service that helps a robot to recognize acoustic patterns in a human's tone of voice by classifying approving, neutral and prohibitive affects. Kim et al. \cite{kim2009people}
identified three phases during which humans use affective prosody during interactive reinforcement learning. These three phases are: direction (before the learner acts), guidance (as the learner indicates intent) and feedback (after the learner completes a task-action).  
Short et al. \cite{short2018detecting} used ratio of variances in raw speech features to detect contingent human responses to robot probes for open-world human-robot interaction. However, in comparison to the problem settings of 
these prior works,
 we focus on the learning from demonstration paradigm 
and characterize raw audio signals of human teachers to better understand the information present in this modality.
Work on infant-directed demonstrations~\cite{nagai2008toward} as well as robotics work inspired from infant-directed speech~\cite{nakamura2015constructing} highlights that the way that people talk to a robot could possibly help it learn more easily if they treat it like it is an infant, than if they spoke to it like it is an adult. However, this may not always be possible with a variety of commercialized robots not exhibiting infant-like appearances or personalities. In our work, we study how human demonstrators use unrestricted speech to teach robots which are not infant-like in appearance or behavior, to better inform how human audio cues could be leveraged for learning by a variety of robots. 

\subsection{Human Audio-assisted Robot Learning}
Prior research in learning from demonstration 
has utilized human speech signals accompanying demonstrations, however with a restricted vocabulary of words used by the human teacher. Nicolescu et al. \cite{nicolescu2003natural} demonstrated the role of verbal cues both during demonstrations and as feedback from the human teacher during the agent's learning process, to facilitate learning of navigation behaviors on a mobile robot. However, they restricted human teachers to use a limited vocabulary of words to indicate relevant parts of the workspace or actions that a robot must execute. Similarly, 
Pardowitz et al. \cite{pardowitz2007incremental} used a fixed set of seven vocal comments which are mapped one-to-one with features relevant to the task 
to augment subtask similarity detection and learning of the task model from demonstrations 
for a simple table setting task. Prior work in reinforcement learning has also utilized a fixed vocabulary of words to record voice-based feedback for reward shaping~\cite{tenorio2010dynamic} and reinforcement learning ~\cite{kim2007learning}. Some works have also utilized sentiment analysis using natural language processing to aid reinforcement learning with the semantics of spoken words \cite{krening2016learning, krening2018newtonian}.
However, in our work, we use raw and unrestricted speech signals accompanying demonstrations to understand how human teachers convey information via audio.
\section{Methodology}

\begin{figure}
\centering
\subfigure[Natural Instruction Type]{
\includegraphics[width=0.225\textwidth]{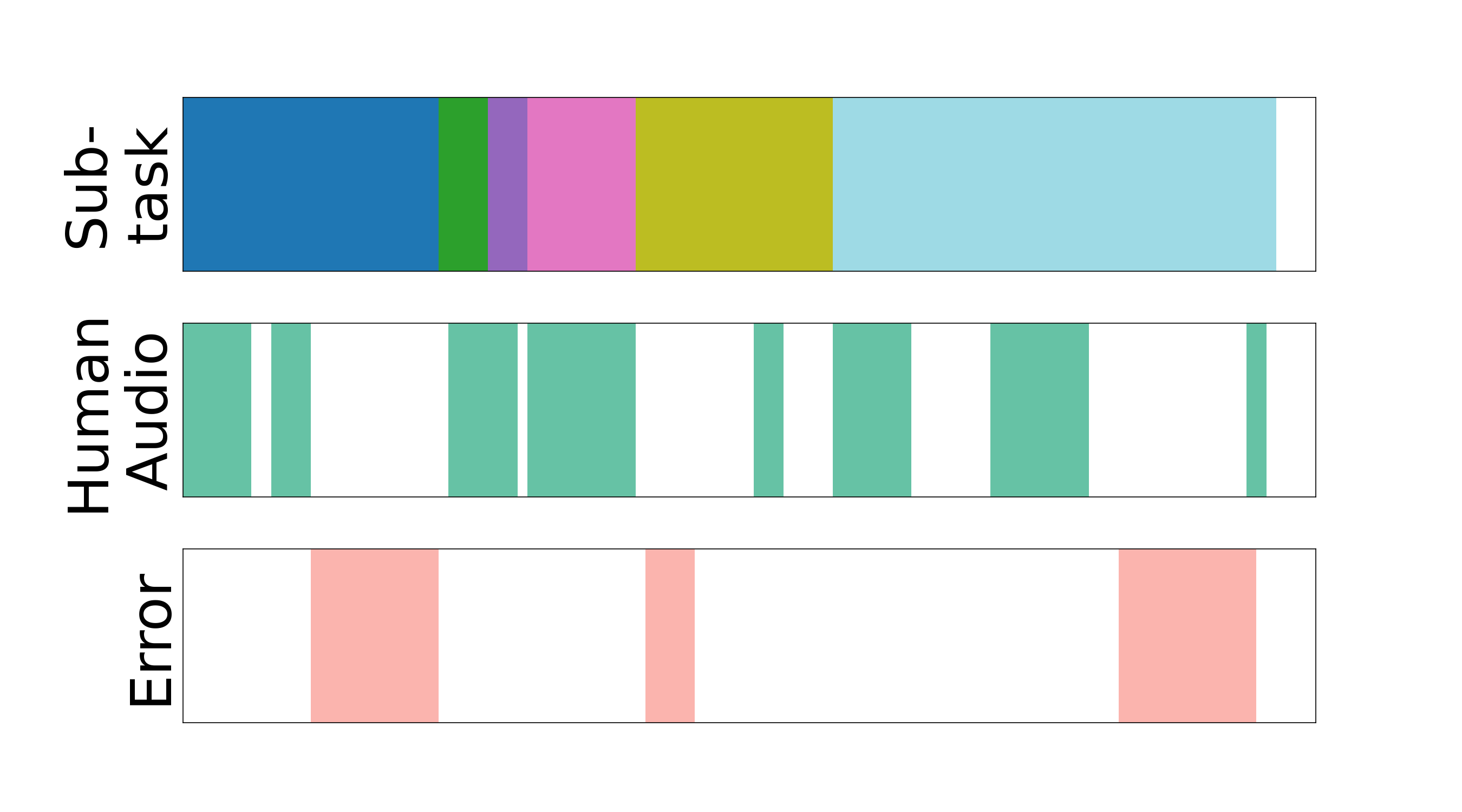}
}
\subfigure[Narration Instruction Type]{
\includegraphics[width=0.225\textwidth]{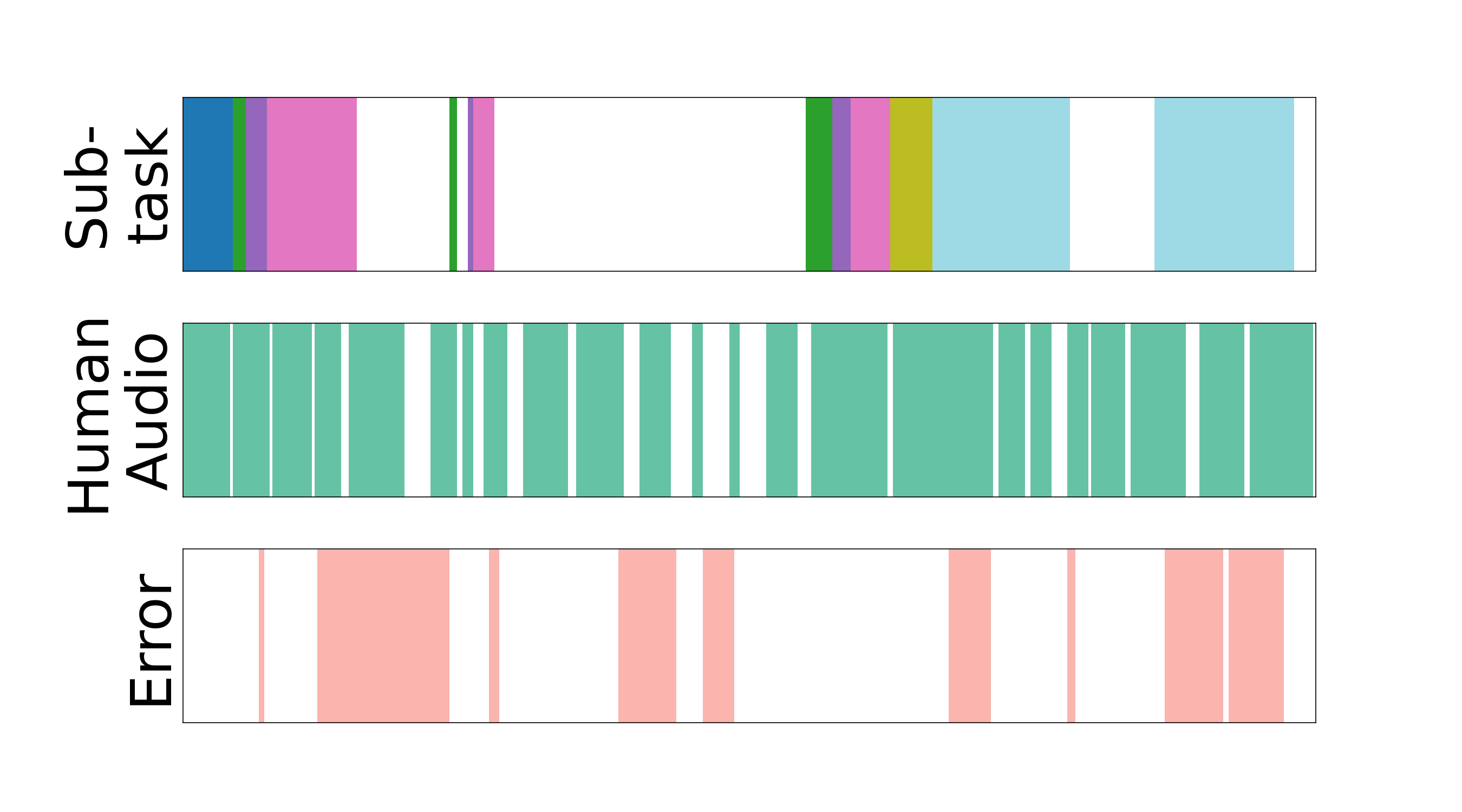}
}

\caption{Subtasks, human  
utterances, and errors during demonstrations are shown on the top, middle, and bottom rows respectively during kinesthetic demonstrations for the fruit cutting task (each subtask category represented with a different color, all error categories represented by the same color). 
}
\vspace{-5mm}
\label{fig:seg_errors}
\end{figure}

To develop an understanding of complex audio signals from human teachers demonstrating manipulation tasks to an embodied robot, we aim to understand how three different acoustic properties (duration, prosody, semantic content) are used by human teachers across demonstration modalities, audio-usage instructions, relevant subtask segments, and errors during demonstrations.
To analyze this information, we conducted a user study (Sec. \ref{sec:user-study}) where human teachers demonstrate two multi-step manipulation tasks (Sec. \ref{sec:tasks}) to a Sawyer robot. 
We recorded audio data
along with visual state observations for two demonstration modalities --- kinesthetic teaching (KT) and video demonstrations (Sec. \ref{sec:system}).
Half of the participants were asked to explicitly narrate what they were doing as they demonstrated the task and the other half were only implictly told that the robot can watch and listen to them (instruction type).
To characterize audio utterances of human teachers accurately, we collected annotations for when teachers talk and what they conveyed through it. To understand how utterances relate to 
task segments and errors during demonstrations, we also collected annotations for (1) subtask segment labels,  
(2) types of demonstration errors, and 
(3) characterizations of human speech accompanying subtasks and errors. 
Details about our annotation procedure are explained in Sec \ref{sec:annotation}.  
Finally, in light of recent LfD methods which (1) highlight that learning a separate policy for each subtask in a multi-step task is more efficient~\cite{kroemer2019review} and (2) utilize suboptimal demonstrations~\cite{osa2018algorithmic}, we analyze if human audio contains beneficial information for two related learning problems --- detecting the presence of subtasks and detecting the presence of demonstration errors. Through proof-of-concept experiments, we demonstrate that random forest classifiers can detect subtasks \& errors during demonstrations (Sec. \ref{sec:learning-method}) with a varied set of human acoustic features (Sec. \ref{sec:features}).

\subsection{Task Descriptions}
\label{sec:tasks}
The two household tasks relevant to personal robots that were used in our study are: (a) box opening and (b) fruit cutting (Fig.~\ref{fig:sawyer-tasks}). 
Each task consists of multiple subtasks performed in sequence to accomplish the overall goal. There could be pauses and errors during and between the execution of the subtasks. Details of the two tasks are as follows:
\subsubsection{Box Opening} A transparent box contains a green wooden object inside it. A circular button located in the center of the lid (covered with black paper) is pressed. This unlocks the lid from the box. The lifted circular part of the lid is then grasped to open the box. Then, the green wooden object is picked from inside the box and placed on the table. The lid is placed on top of the box. The circular button of the lid is pressed again to lock the box. Finally, the green object is picked from the table and placed on top of the black circular button of the lid. The goal is to perform the task without moving the box from its location on the table surface (Fig.~\ref{fig:sawyer-tasks}(a)). This is a multi-step pick-and-place task with a challenging maneuver of lifting an object from inside the box. This task is representative of several goal-oriented household tasks.

\subsubsection{Fruit Cutting}
A flat surface contains different colored wooden fruit pieces on it, along with an uncut wooden banana whose three pieces are stuck together with Velcro. A wooden knife is placed in a knife holder. The gripper of the knife is covered with foam so it can be easily grasped by the robot.
A slit with a black marking indicates where the robot's gripper can firmly grasp the knife. The knife is taken out of the knife holder, and then used to cut the banana into three pieces by making two cuts through the two Velcro attachments. Finally, the knife is placed back into the knife holder. The goal is to cut the banana without moving any other fruit on the wooden surface such that the cut pieces also do not fall off (Fig.~\ref{fig:sawyer-tasks}(b)). This is a task which requires complex maneuvers to successfully cut the fruit pieces. This is representative of complex manipulation tasks where the trajectory path and forces applied can matter for successfully achieving the goal.

\begin{table}[htb]
    \centering
    \caption{Ordered subtask categories
    of the box opening task.}
    \begin{tabular}{|c|c|}
    \hline
    Index & Subtask Category\\
    \hline\hline
    1 & Move arm towards the lid on the closed box\\
    \hline
    2 & Click button to open the lid of the box \\
    \hline
    3 & Grasp button on the lid of the box \\
    \hline
    4 & Transport the lid from the box towards table\\
    \hline
    5 & Release the button of the lid on the table from the gripper\\
    \hline
    6 & Move arm towards the green object that's inside the box\\
    \hline
    7 & Grasp the green object\\
    \hline
    8 & \specialcell{Transport the green object from\\ inside the box towards the table}\\
    \hline
    9 & Release the green object from gripper\\
    \hline
    10 & Move arm towards the lid on the table\\
    \hline
    11 & Grasp the button of the lid placed on the table\\
    \hline
    12 & Transport the lid from the table towards the box's body\\
    \hline
    13 & \specialcell{Release the button of the lid\\ on top of the box from the gripper}\\
    \hline
    14 & Push the button to lock lid\\
    \hline
    15 & Move arm towards green object\\
    \hline
    16 & Grasp green object on the table\\
    \hline
    17 & Transport the green object towards lid button \\
    \hline
    18 & Release the green object from gripper on top of the lid\\
    \hline
    \end{tabular}
    \label{tab:box_seg_labels}
\end{table}

\subsection{Robot System}
\label{sec:system}

Demonstrations were provided to a 7 degree-of-freedom Sawyer manipulator with series elastic actuators and a parallel gripper. We used
a Movo MC1000 Conference USB microphone to record audio during demonstrations. All  participants were told that the robot can watch and listen to them. 
We used three static camera sensors (facing the human demonstrator and workspace) to collect high resolution video
data at $\sim\hspace{-1.5mm}10$ Hz for observing the demonstration. 
The recorded visual and audio data were synchronized and collected using ROS~\cite{quigley2009ros}.

\begin{table}[htb]
    \centering
    \caption{Ordered subtask categories
    of the fruit cutting task.}
    \begin{tabular}{|c|c|}
    \hline
    Index & Subtask Category\\
    \hline\hline
    1 & Move arm towards the knife\\
    \hline
    2 & Grasp the knife while it's inside the knife holder\\
    \hline
    3 & Move the knife out from knife holder \\
    \hline
    4 & \specialcell{Transport knife from the \\knife holder towards the banana}\\
    \hline
    5 & Cut the first piece of the banana\\
    \hline
    6 & Cut the second piece of the banana\\
    \hline
    7 & \specialcell{Transport knife back from the\\ cutting board to the knife holder}\\
    \hline
    8 & Place the knife back in the knife holder\\
    \hline
    9 & Release the knife from the gripper\\
    \hline
    \end{tabular}
    \label{tab:cutting_seg_labels}
\end{table}

\subsection{User Study}
\label{sec:user-study}
\subsubsection{Independent Variables (IVs)}
\label{sec:IVs}
We evaluated the effects of two independent variables (demonstration type and instruction type) on different acoustic features. For demonstration type (within-subjects), we focus on two modalities of LfD for robot manipulation \cite{kroemer2019review}: (1) learning via kinesthetic teaching (KT) in which the joints of a robot are moved along a trajectory in order to accomplish the task \cite{akgun2012trajectories}, and (2) learning from observation, specifically video demonstrations, in which a robot passively observes a human performing the task. For instruction type (between-subjects), we focused on the narration and natural conditions. Half of the participants were part of the narration condition, in which they were explicitly asked to use speech to communicate their intentions as they demonstrate. The remaining half of the participants were in the natural condition, in which they were only told that the robot can listen to them but not asked to use speech specifically. To analyze data for subtasks and errors during demonstrations, an IV indicating their presence or absence is used (example of presence/absence depicted in Fig.~\ref{fig:seg_errors}).

\subsubsection{Data Collection Procedure}
We collected demonstration data from 20 participants (8 males, 12 females) for each of the two tasks (Fig. \ref{fig:sawyer-tasks}). Participants were graduate or undergraduate students, recruited from a university campus. All participants used English speech and were 
proficient at speaking the language. Each participant was 
allowed one practice round for each demonstration type on the task they were assigned to do first. After one round of practicing, participants completed 2 demonstrations (1 KT, 1 video) for the box opening task and 2 demonstrations (1 KT, 1 video) for the fruit cutting task. The order of tasks and demonstration types were counterbalanced across all participants. 
We discarded data from two participants due to network issues during recording and perform our analysis on the remaining 18 participants. This amounted to a total of $\sim$24 minutes of video demonstration data and $\sim$119 minutes of KT demonstration data.

\subsubsection{Dependent Measures (DMs)}
\label{sec:dv}
Our dependent measures are the (1) duration 
of utterances during demonstrations (also measured as speech density, i.e. fraction of a demonstration accompanied by speech), (2) acoustic features capturing the prosody of speech, and (3) richness of the content conveyed by 
utterances. Richness of the content refers to the diversity of the concepts conveyed via the spoken words. 

\subsubsection{Research Hypotheses}
Our research hypotheses are as follows:
\begin{itemize}
    \item \textbf{H1}: Prior work has shown that humans rely more on additional channels of communicating intent, such as audio, during challenging tasks \cite{oviatt2004we}. Based on this finding and given the complexity of our manipulation tasks, we hypothesize that teachers would rely on using  
    similar amount (speech density) and kind of
    utterances (expressiveness or prosody) under both the natural and narration instruction conditions.
    \item \textbf{H2}: Following the previous motivation from \cite{oviatt2004we}---since kinesthetic demonstrations require more physical effort from the human teacher compared to video demonstrations \cite{fischer2016comparison}, 
    we hypothesize that more pronounced acoustic features and higher density of speech would be present during kinesthetic demonstrations.
    \item \textbf{H3}: 
    Prior work which studied unrestricted audio signals in human tutoring from parents to children 
    \cite{schillingmann2009computational}, established that human speech from the demonstrator binds to action events, with structured pauses between events.
    Given these findings, we hypothesize that human teachers' speech 
    would be more pronounced in terms of density and prosodic features \textit{during} relevant subtask executions in a demonstration (presence), compared to periods of gaps between subtasks (absence). 
    \item \textbf{H4}: Human 
    utterances have been shown to be beneficial for improving the error prediction of ASR systems \cite{hirschberg2004prosodic}, with more emphatic uses of speech when errors occur. Thus, we hypothesize that human teachers' speech would be more expressive and emphatic during presence of error segments versus in their absence.
\end{itemize}

\subsection{Data Annotation} 
\label{sec:annotation}

Automated speech detection and recognition is an active and challenging research area \cite{yu2016automatic, juang2005automatic, benzeghiba2007automatic}.
Average word error rates on our dataset (computed using the publicly available Google Speech to Text API~\cite{googlestt}) are 0.37 (video demos) and 0.59 (KT demos).
To accurately characterize human speech, subtasks, and errors during demonstrations, we collected detailed annotations from three different human annotators. 
All annotators were undergraduate students recruited at a university campus. The following annotations were collected for each utterance (separated by a significant pause, as judged by the annotators.): start time, stop time, and speech transcription. 
Each annotation was provided via a GUI interface 
with information available from the audio microphone and all three camera views to be played back at any time. 

\begin{table}[htb]
    \centering
    \caption{Unordered error categories
    of the box opening and fruit cutting tasks.}
    \begin{tabular}{|c|c|}
    \hline
    Index & Error Category\\
    \hline\hline
    1 & \specialcell{Teacher forgets to perform\\ a step of the task}\\
    \hline
    2 & \specialcell{Teacher struggles to move the\\ robot's arm in a certain way (KT only)}\\
    \hline
    3 & \specialcell{Teacher struggles to grasp an object} 
    \\
    \hline
    4 & \specialcell{Robot/Human arm collides with, knocks off,\\ or moves items from their original position}\\
    \hline
    5 & \specialcell{Teacher performs a step of\\ the task in the wrong order}\\
    \hline
    6 & \specialcell{Teacher unintentionally uses two hands\\ instead of one to perform the task (video only)}\\
    \hline
    7 & \specialcell{Teacher uses their human hand during a kinesthetic\\demonstration to complete the task/help the robot (KT only)}\\
    \hline
    8 & \specialcell{Teacher accidentally drops an object\\ already grasped by the robot's gripper (KT only)}\\
    \hline
    9 & \specialcell{Teacher intentionally re-strategizes\\ or re-attempts a step of the task} \\
    \hline
    10 & \specialcell{Unsuccessful step execution e.g. not applying\\ enough force with the knife to completely cut the banana}
    \\
    \hline
    11 & Other\\
    \hline
    \end{tabular}
    \label{tab:err_labels}
\end{table}

Annotators also separately marked the start and end times of predefined error and 
subtask categories for each demonstration (Fig. \ref{fig:seg_errors}), along with labels of predefined speech categories present with each identified subtask and error. 
The predefined subtask categories 
are listed in Table~\ref{tab:box_seg_labels} for the box opening task 
and in Table~\ref{tab:cutting_seg_labels} for the fruit cutting task. 
Each subtask category represents a primitive step that a robot can learn a policy for. 
The predefined error categories for both tasks are listed in Table~\ref{tab:err_labels}.  
The predefined speech categories that were labeled along with subtask and error instances were: (1) frustration, (2) encouragement, (3) speech pauses (explicit pause or filler words between two speech utterances), (4) laughter, (5) surprise, (6) no variation in manner of speech (normal). Annotators could also choose to label an utterance in a `none' category if it did fit well under any of the predefined categories. All predefined categories were determined following the methodology of grounded theory analysis \cite{charmaz2007grounded,charmaz2014constructing,morse2016developing}.

\subsection{Acoustic Feature Computation}
\label{sec:features}

We compute several acoustic features over the annotated utterances to characterize prosody and content of human speech.  
We first processed raw audio signals of human demonstrators with a speech enhancement model~\cite{defossez2020real} to filter out the effect of environmental object-interaction sounds and robot motor noises. We then compute 10 
hand-crafted prosodic features (Audio II) 
to capture prosody and affect in terms of variation in the loudness or speed with which a speaker can alter their pronouncement of an utterance \cite{hirschberg2004prosodic,tran2017parsing,kim2007learning,short2018detecting,kim2018crepe} 
--- 
maximum pitch, energy, and loudness; mean pitch, energy, and loudness; 
total energy;
pause duration between speech acts; total word count and word rate. Pitch, loudness, and energy features represent measures of intonation and enunciation, whereas duration, pause, word count, word rate capture timing information. 
These hand-crafted acoustic features have been shown to enhance semantic parsing \cite{tran2017parsing}, understand speech recognition failures in dialogue systems \cite{hirschberg2004prosodic}, and widely used for applications in  human-robot interaction~\cite{kim2009people,short2018detecting} and speech recognition~\cite{yu2016automatic} communities. 

\begin{table}
    \centering
    \caption{Percentage of a demonstration duration accompanied by utterances (speech density), subtasks, and errors. Aggregate statistics (mean \& standard error) are reported for each of the four subsets resulting from two IVs. 
    }
    \begin{tabular}{|c|c|c|c|}
    \hline
    \begin{tabular}[c]{@{}c@{}}Demo Type/\\Instruction Type\end{tabular} & \%Utterances & \%Subtasks & \%Errors \\
    \hline\hline
    Video/Natural  & $57.97 \pm 14.12$ & $79.56 \pm 1.80$  & $9.09 \pm 3.27$  \\
    \hline
    Video/Narration & $89.89 \pm 3.26$  & $90.67 \pm 2.28$  & $4.84 \pm 1.87$   \\
    \hline
    KT/Natural & $27.90 \pm 11.86$ & $91.43 \pm 2.41$  & $20.01 \pm 3.87$  \\
    \hline
    KT/Narration & $71.60 \pm 6.09$  & $85.33 \pm 2.95$  & $24.57 \pm 2.74$ \\
    \hline
    \end{tabular}
    \label{tab:stats}
\end{table}

We also analyze annotated prosodic features (Audio I) in the form of  emotion labels present in speech 
(Sec.~\ref{sec:annotation}). In addition to the annotated (Audio I) and hand-crafted acoustic features (Audio II), we also compute 256-dimensional deep acoustic features (PASE~\cite{pascual2019learning}) for our learning experiments (Sec.~\ref{sec:learning-method}). PASE features are obtained from the last layer of a problem-agnostic speech encoder, trained in a self-supervised manner, and beneficial for downstream tasks which leverage speech features. 
To understand the richness of semantic content conveyed via audio utterances, we used variance of GloVe embeddings \cite{pennington2014glove} (first principal component from PCA analysis) for words in the annotated speech transcriptions.  
GloVe embeddings are word representations that capture fine-grained semantic and syntactic regularities of words using vector arithmetic.
The variance of such embeddings 
can quantify the variety of concepts communicated.

\subsection{Learning Classifiers for Subtask and Error Detection}
\label{sec:learning-method}

To further understand if different acoustic features in human speech can identify subtasks and errors during demonstrations, we use random forest classifiers ~\cite{pedregosa2011scikit} for two binary classification tasks: (1) subtask detection and (2) error detection. 
The default implementation from the scikit-learn Python library~\cite{pedregosa2011scikit} is used for our experiments.
We use $100$ trees for each experiment. From the human facing camera, we also sample every $16^{th}$ frame
to compute video features (penultimate layer output of I3D~\cite{carreira2017quo} pretrained on the Kinetics activity recognition dataset~\cite{kay2017kinetics}) and corresponding audio features (Sec.~\ref{sec:features}) for a 1 second window around this frame. 
This provides about $2000$ (cutting) and
$2500$ (box) samples for kinesthetic demonstrations; and $370$ (cutting) and 
$510$ (box) samples for video demonstrations. Audio features consist of annotated speech labels encoded as one-hot vectors (Audio I) described in Sec.~\ref{sec:annotation}, hand-crafted prosodic features (Audio II) described in Sec.~\ref{sec:features}, as well as deep acoustic features (PASE~\cite{pascual2019learning}). 

\section{Results and Discussion} 
We tested the reliability across annotators for consistency of timing and content they labeled. The Interclass Correlation (ICC) test shows high agreement on timing information between the annotators for  
utterances, segments and mistakes with $ICC(1)=0.94, F(35,10)=41.6, p<0.001$ (utterances); $ICC(1)=0.82, F(971,971)=10.1, p<0.001$ (segments); $ICC(1)=0.89, F(35,33)=18.4, p<0.001$ (mistakes). Annotations for the error categories, subtask categories, and speech categories also have good reliability with Cohen's $\kappa > 0.72$. Due to the high agreement among annotators, we used data from a single annotator to present the results going forward. 

We report various findings about utterances (Sec.~\ref{sec:dur}), subtasks and errors (Sec.~\ref{sec:err_seg}), and along the way address if our hypotheses are supported by the results. Unless otherwise noted, a statistical model based on a 2 x 2 mixed design with instruction type as the between-subjects factor and demonstration type as the within-subjects factor was used in the analyses of variance (ANOVA).  
We also report findings from our learning experiments in Sec.~\ref{sec:learning-results}. 

\begin{table}
    \centering
    \caption{Means and standard errors of annotated acoustic feature density (\%) (Audio I) computed across entire demonstrations.
    }
    \resizebox{0.48\textwidth}{!}{
    \begin{tabular}{|c||c|c|c|c|c|c|}
    \hline
 & \specialcell{Frustration} & 
 \specialcell{Surprise}&
 \specialcell{Speech\\Pauses} &   \specialcell{Normal\\Speech} &
 \specialcell{Laughter}  & \specialcell{Encourage\\-ment}\\
\hline
KT        & 2.07$\pm$1.03 &	3.67$\pm$1.39 &	8.31$\pm$2.23	&65.95$\pm$6.92	&1.01$\pm$1.03	&2.94$\pm$1.62  \\
\hline
Video     &  0.00$\pm$0.00 &	0.10$\pm$0.10	&1.80$\pm$1.33	&62.81$\pm$7.60	&0.00$\pm$0.00	&0.00$\pm$0.00 \\
\hline
F(1,16)   &  5.01	&8.59	&13.08	&0.63	&1.00	&3.74 \\
\hline
p         &  \textbf{\textless{}0.05} &	\textbf{\textless{}0.01}	&\textbf{\textless{}0.01}	&0.44	&0.33	&0.07 \\
\hline \hline
Narration   &  2.07$\pm$1.03	&3.37$\pm$1.40	&2.17$\pm$1.26	&81.22$\pm$3.01	&1.01$\pm$1.01	&2.83$\pm$1.62\\
\hline
Natural &  0.00$\pm$0.00	&0.40$\pm$0.34	&7.95$\pm$2.33	&47.54$\pm$7.97	&0.00$\pm$0.00	&0.11$\pm$0.11\\
\hline
F(1,16)   & 5.01	&5.46	&3.84	&8.23	&1.00	&3.26\\
\hline
p         & \textbf{\textless{}0.05}	&\textbf{\textless{}0.05}	&0.07	&\textbf{\textless{}0.05}	&0.33	&0.09\\
\hline
\end{tabular}}
\label{tab:stats_annot}
\end{table}

\subsection{Human Audio Analysis during Demonstrations (\textbf{H1}, \textbf{H2})}
\label{sec:dur}

\begin{table*}
    \centering
    \caption{Means and standard errors of speech density, hand-crafted acoustic features (Audio II), and Variance of GloVe embeddings computed across entire demonstrations.
    }
    \resizebox{\textwidth}{!}{
    \begin{tabular}{|c||c||c|c|c|c|c|c|c|c|c|c||c|}
    \hline
 & \specialcell{Speech\\Density (\%)} & \specialcell{Mean\\Pitch} & \specialcell{Max\\Pitch} & \specialcell{Mean\\Energy(1E-4)}  & \specialcell{Max\\Energy} & \specialcell{Total\\Energy} & \specialcell{Mean\\Loudness(1E-4)} & \specialcell{Max\\Loudness} &  \specialcell{Word\\Count} & \specialcell{Word\\Rate} &
 \specialcell{Pause\\Density (\%)}&
 \specialcell{GloVe \\(PC Variance)}  \\
\hline
KT      & $48.61\pm7.98$   & $120.18\pm14.61$ & $264.48\pm41.56$ & $2.92\pm0.58$    & $0.21\pm0.07$    & $983.52\pm305.05$ &  $56.50\pm9.06$  & $0.37\pm0.07$    & $154.81\pm40.97$ & $1.10\pm0.16$     & $25.69\pm3.99$    & $2.85 \pm 0.39$ \\
\hline
Video   & $64.76\pm8.68$   & $121.18\pm17.9$  & $227.33\pm39.73$ & $3.65\pm0.69$    & $0.08\pm0.02$    & $374.50\pm138.83$  &  $73.09\pm11.45$  & $0.24\pm0.04$    & $68.78\pm21.44$ & $1.75\pm0.25$    & $17.62\pm4.34$    & $2.77\pm0.39$ \\
\hline
F (1,16) & $10.86$ & $0.03$ & $3.75$ & $6.91$ & $5.40$  & $7.15$ & $19.39$ & $8.62$ & $8.52$ & $45.21$ & $10.86$ & $0.28$ \\
\hline
p       & $\textbf{\textless{}0.01}$ & $0.87$            & $0.07$            & $\textbf{\textless{}0.05}$ & $\textbf{\textless{}0.05}$ & $\textbf{\textless{}0.05}$  & $\textbf{\textless{}0.01}$ & $\textbf{\textless{}0.01}$ & $\textbf{\textless{}0.05}$ & $\textbf{\textless{}0.01}$ & $\textbf{\textless{}0.01}$ & $0.60$ \\
\hline \hline
Narration & $76.02\pm16.49$ & $145.16\pm8.61$ & $312.25\pm32.93$ & $3.95\pm0.55$ & $0.23\pm0.06$ & $941.22\pm281.56$ & $77.69\pm7.00$ & $0.43\pm0.06$ & $160.53\pm167.14$ & $1.73\pm0.15$ & $11.99\pm8.25$ & $3.49\pm0.1$  \\
\hline
Natural  & $37.35\pm39.70$   & $96.21\pm19.73$ & $179.56\pm41.76$ & $2.62\pm0.69$ & $0.06\pm0.02$ & $416.80\pm189.4$ & $51.91\pm12.36$ & $0.19\pm0.04$ & $63.06\pm97.12$ & $1.11\pm0.26$ & $31.33\pm19.85$ & $2.13\pm0.49$ \\
\hline
F (1,16) & $8.41$ & $2.54$ & $3.18$ & $1.20$  & $7.81$ & $1.65$ & $1.73$ & $9.87$ & $3.23$ & $2.69$ & $8.41$ & $3.62$ \\
\hline
p & $\textbf{\textless{}0.05}$ & $0.13$ & $0.09$ & $0.29$ & $0.01$ & $0.22$ & $0.21$ & $\textbf{\textless{}0.01}$ & $0.09$ & $0.12$ & $\textbf{\textless{}0.05}$ & $0.08$  \\                    

\hline
\end{tabular}}
\label{tab:stats_prosody}
\end{table*}

\subsubsection{Quantification of 
Utterances}
\label{sec:timing}

The total duration of the demonstrations are $4069.37$ sec (Narration/KT), $3080.34$ sec (Natural/KT), $893.95$ sec (Narration/Video),
$521.86$ (Natural/Video).
The total human audio duration are $2910.15$ sec (Narration/KT), $885.30$ sec (Natural/KT), $811.11$ sec (Narration/Video), $306.56$ sec (Natural/Video).
The ANOVA results reveal that the percent duration of a demonstration 
accompanied by utterances (speech density) 
has a significant main effect along both IVs: instruction type ($F(1,16)=8.41,p<0.05$) and demonstration type ($F(1,16)=10.86,p<0.01$) (column 2 of Table~\ref{tab:stats}). 
Speech density is significantly more in the narration instruction conditions ($M=76.02, SD=16.49$) in comparison to the natural instruction conditions ($M=37.35, SD=39.70$). Thus, the results for speech density do not provide support for H1.
Speech density is significantly more during video demonstrations ($M=64.76, SD=36.81$) compared to KT demonstrations ($M=48.61, SD=33.87$). This result can be explained by the fact that there are longer pauses between utterances (normalized by demonstration duration) during KT ($M=25.69\%, SD=16.93$) compared to video ($M=17.62\%, SD=18.40$) demonstrations ($F(1,16)=26.89, p<0.01$). Often users struggle with moving the robot arm in the right configuration during KT demos, and focus on executing subtasks instead of simultaneously talking during such periods. Thus, speech density does not provide support for H2.

\subsubsection{Speech Prosody}
\label{sec:audio_prosody}
The annotated (Audio I) and hand-crafted (Audio II) prosodic feature values are listed in Table~\ref{tab:stats_annot} and columns 3-12 of Table \ref{tab:stats_prosody} respectively. The feature values are accumulated 
for 
utterances in a single demonstration and these accumulated values are averaged across demonstrations. 
For Audio I features (Table~\ref{tab:stats_annot}), we observe that most demonstrations are accompanied with normal speech (without much expressivity via emotions). 
Frustration ($F(1,16)=5.01, p<0.05$), surprise ($F(1,16)=8.59, p<0.01$), and speech pauses ($F(1,16)=13.08, p<0.01$) are conveyed significantly more during KT versus video demos.
Since more errors occur during KT demos (column 4 of Table~\ref{tab:stats}, Sec.~\ref{sec:prosody_err_seg}), the presence of frustration, surprise, and pauses also indicated that most errors are unintentional as the users maneuver the robot arm (often getting stuck in singular configurations) and the light-weight gripper (objects often fall from the grasp of the gripper). This finding partially supports H2. 
Frustration ($F(1,16)=5.01, p<0.05$), surprise ($F(1,16)=3.37, p<0.05$), and normal speech ($F(1,16)=8.23, p<0.05$) are conveyed significantly more during narration versus the natural instruction condition. This implies that when users are asked to narrate, they tend to be more expressive, but otherwise do not express their frustration or surprise as emphatically, even during errors. Thus, this finding does not support H1.

For Audio II features (Table~\ref{tab:stats_prosody}), we find only max loudness ($F(1,16)=9.87,p<0.01$) and pause density ($F(1,16)=8.41, p<0.05$) to be significantly higher  for the narration  
versus natural 
condition. The other 8 features are not significantly different across instruction types, thus providing partial support for H1. For demonstration types, KT comprised of more errors and pauses compared to video demos. Thus, the behavior and speech patterns of teachers are different across demonstration types. 8 out of 10 features are significantly higher for KT versus video demos (mean energy, max energy, total energy, mean loudness, max loudness, word count, word rate, pause density), providing partial support for H2.

\subsubsection{Information Conveyed via Spoken Words}

\begin{figure*}
\centering
\subfigure[Video Demonstration (Box Opening Task)]{
\includegraphics[width=0.45\textwidth]{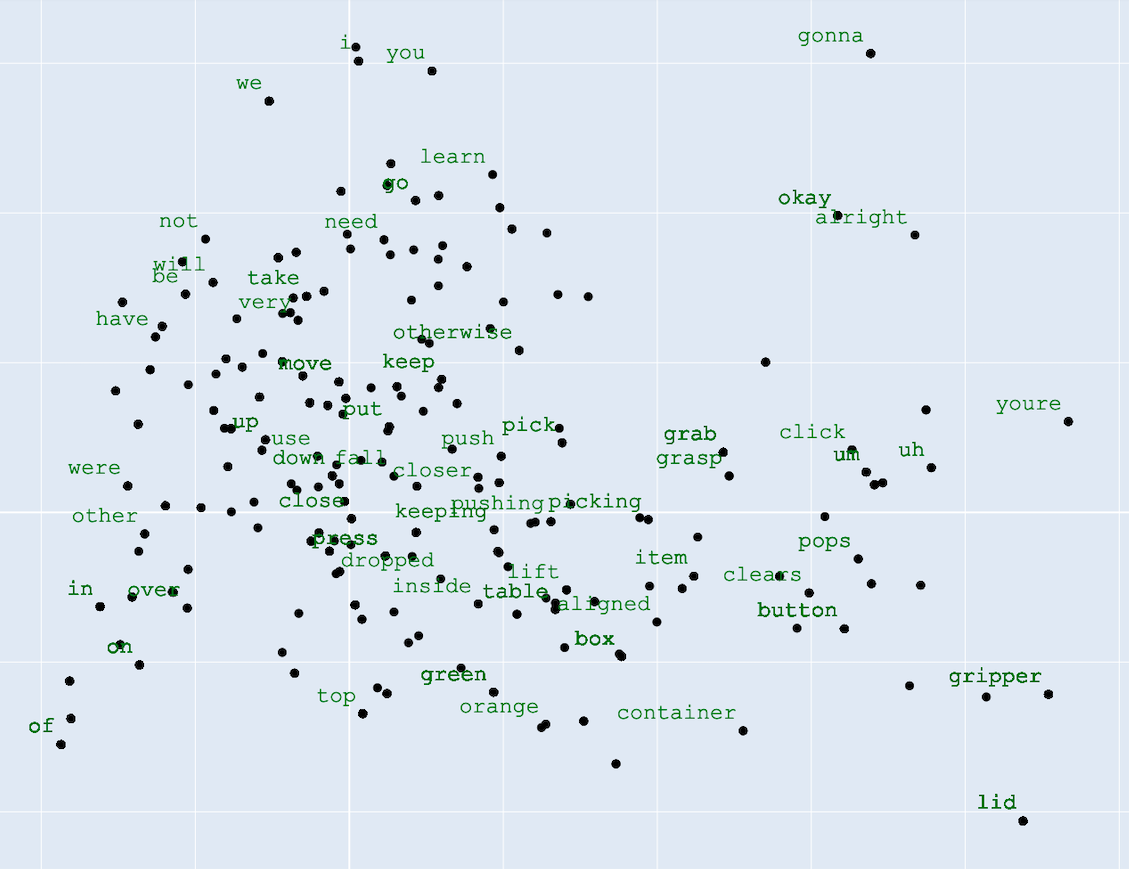}
}
\subfigure[Kinesthetic Demonstration (Box Opening Task)]{
\includegraphics[width=0.45\textwidth]{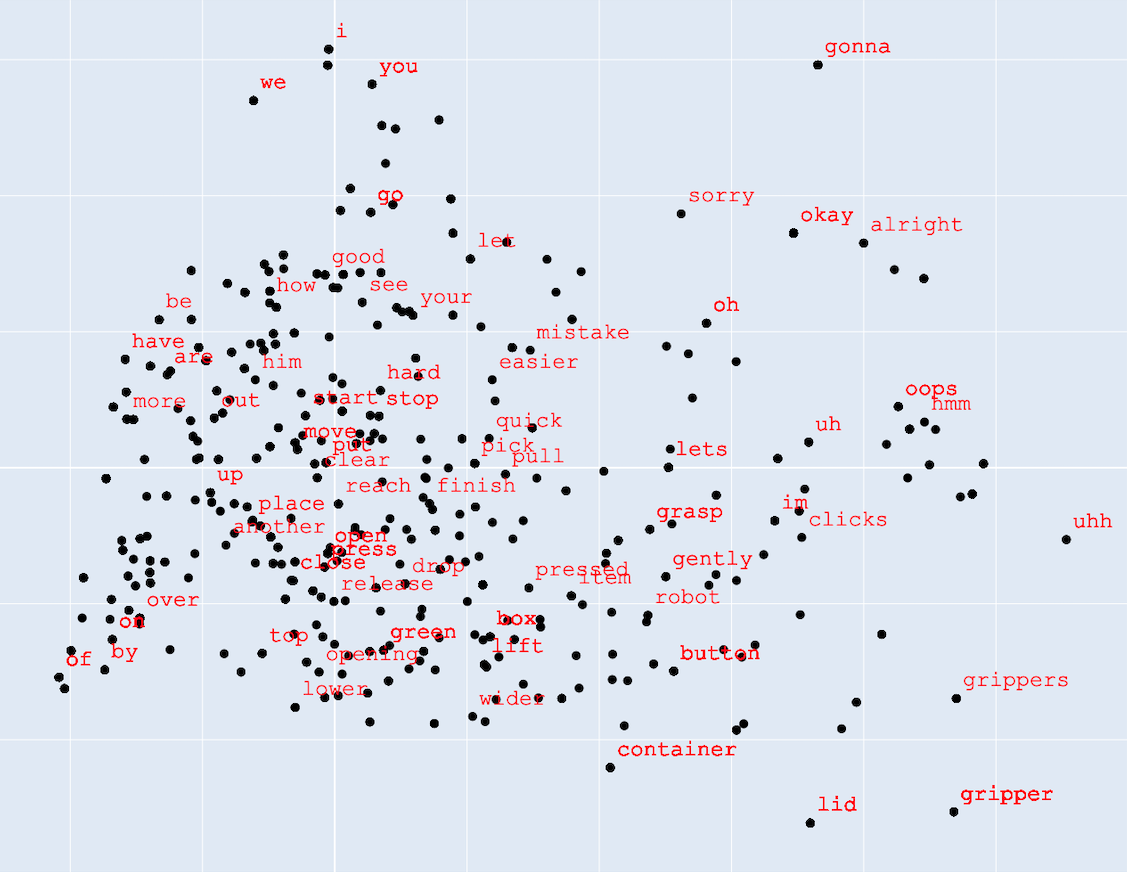}
}
\subfigure[Video Demonstration (Fruit Cutting Task)]{
\includegraphics[width=0.45\textwidth]{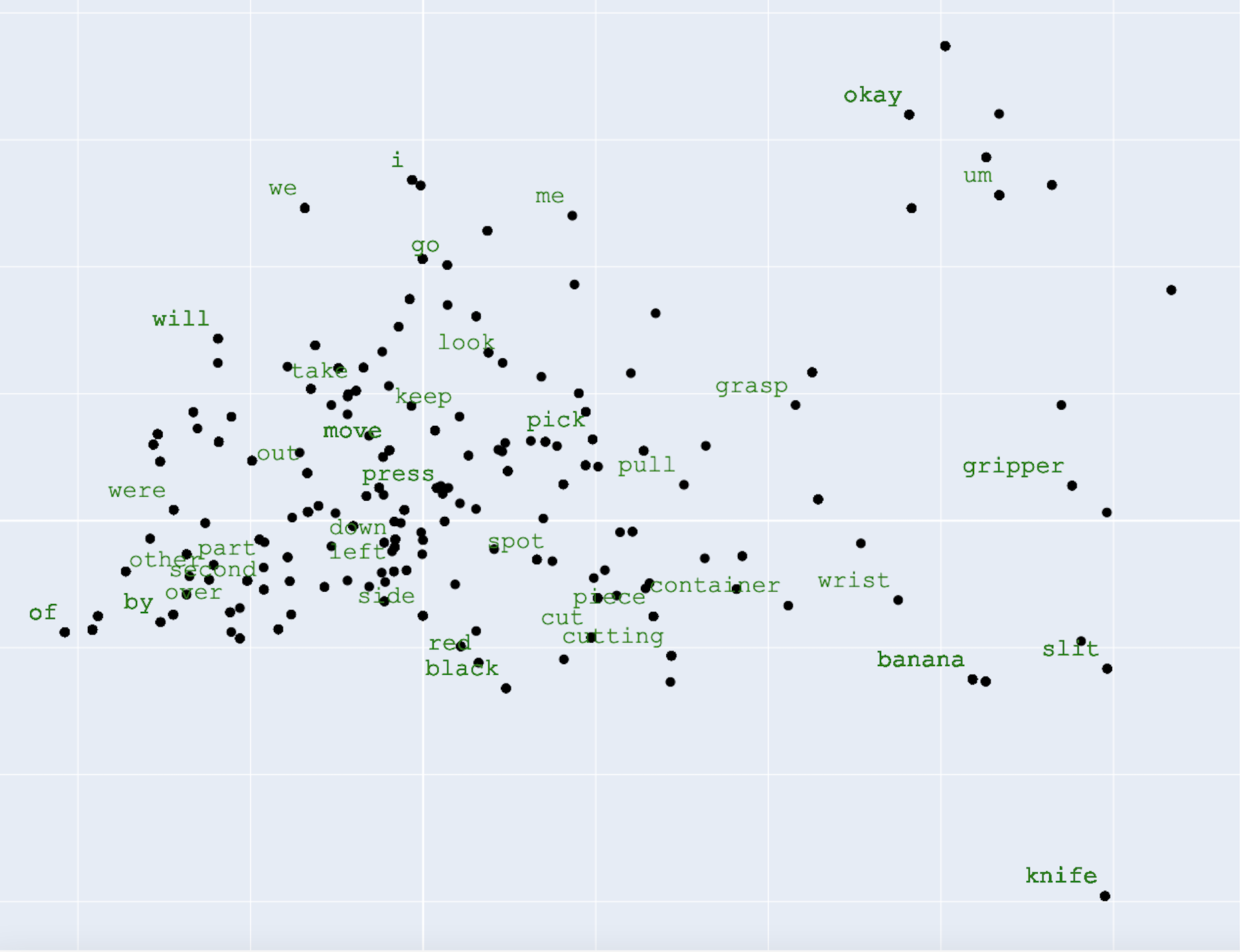}
}
\subfigure[Kinesthetic Demonstration (Fruit Cutting Task)]{
\includegraphics[width=0.45\textwidth]{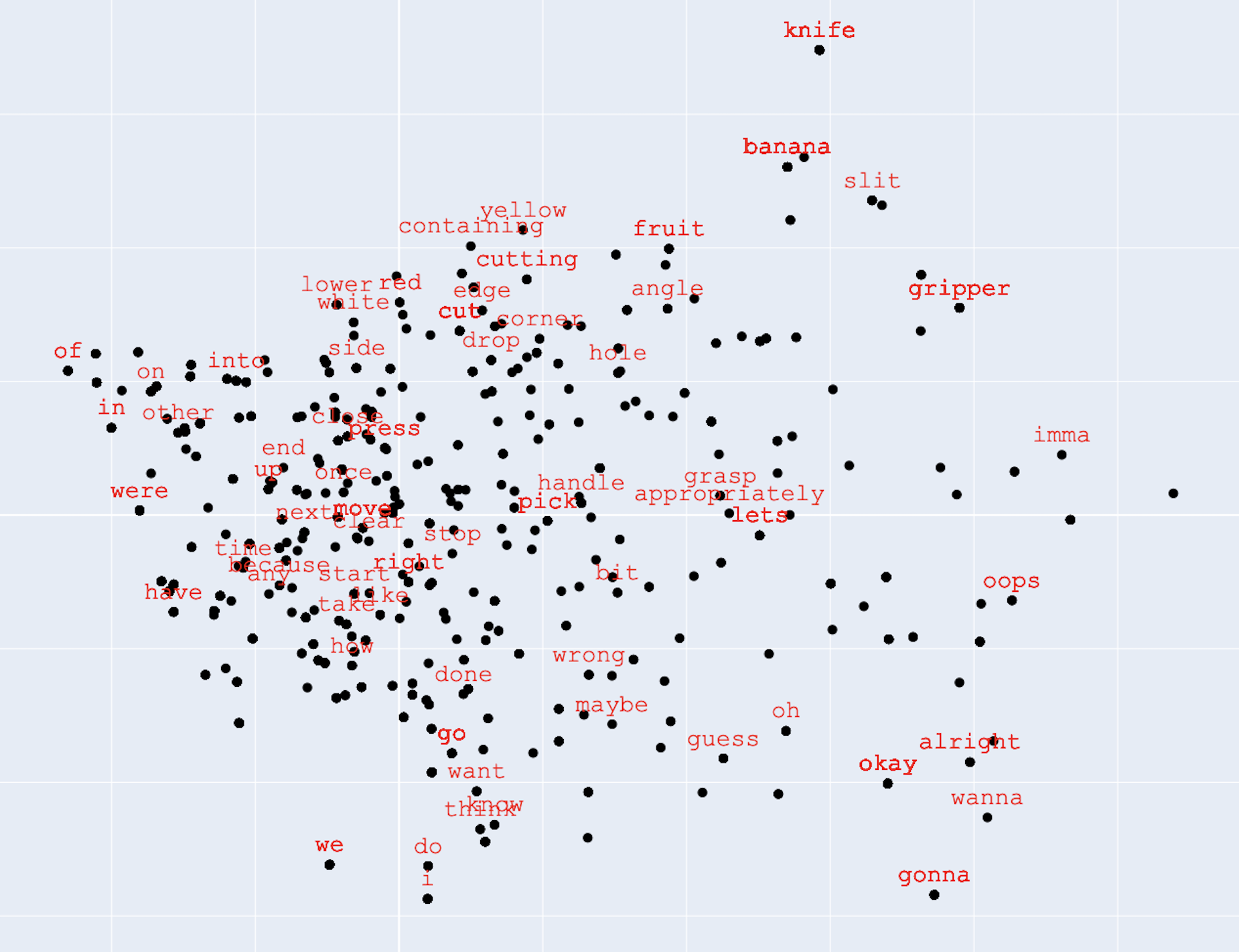}
}
\caption{PCA projections of GloVe word embeddings for speech accompanying demonstrations from all participants. Human teachers used semantically similar words for both kinesthetic and video demonstrations, though a larger vocabulary of words was used during the longer and more challenging kinesthetic demonstrations. 
}
\label{fig:glove}
\end{figure*}

We analyzed the word distributions used in teachers' 
utterances via variance of GloVe vectors across the first principal component (Table~\ref{tab:stats_prosody}). 
The semantic concepts conveyed via spoken words are not significantly different for either demonstration type ($F(1,16)=0.28, p=0.60$) or instruction type ($F(1,16)=3.62, p=0.08$). PCA projections of GloVe word embeddings for the box opening and fruit cutting tasks are shown in Fig. \ref{fig:glove}.  
The overall word count for KT 
is higher than video 
demos ($F(1,16)=8.52,p<0.05$) as shown in Table~\ref{tab:stats_prosody}, with a lot more paraphrasing, prepositions, gerunds, and noun modifiers, and words specific to robot parts (such as `closing', `closed`, `kind of', `picking', `picked', `grip', `gripper',`grasp',`keyframe' etc.) compared to video demos.

\begin{table*}
    \centering
    \caption{Means and standard errors of speech density, hand-crafted acoustic features (Audio II), GloVe embeddings in the presence and absence of subtasks.
    }
    \resizebox{\textwidth}{!}{
    \begin{tabular}{|c||c||c|c|c|c|c|c|c|c|c|c||c|}
    \hline
 & \specialcell{Speech\\Density (\%)} & \specialcell{Mean\\Pitch} & \specialcell{Max\\Pitch} & \specialcell{Mean\\Energy(1E-4)}  & \specialcell{Max\\Energy} & \specialcell{Total\\Energy} & \specialcell{Mean\\Loudness(1E-4)} & \specialcell{Max\\Loudness} &  \specialcell{Word\\Count} & \specialcell{Word\\Rate} &
 \specialcell{Pause\\Density (\%)}&
 \specialcell{GloVe \\(PC Variance)}  \\
 \hline
\textbf{KT/Narration}  &   &  &  &    &    &  &    &     &  &     &    &  \\ 
\hline
Presence      & 68.24$\pm$5.55	 &141.21$\pm$11.90 &	321.51$\pm$43.73 &	3.04$\pm$0.60 &	0.23$\pm$0.06 &	1163.38$\pm$419.53 &	61.78$\pm$7.29 &	0.44$\pm$0.06 &	195.44$\pm$51.91 &	1.29$\pm$0.13	 &31.16$\pm$5.14 &	3.52$\pm$0.20	 \\
\hline
Absence   & 62.10$\pm$7.70	 &145.17$\pm$11.61 &	277.86$\pm$37.23 &	3.41$\pm$0.99&	0.17$\pm$0.10	 &193.24$\pm$79.92 &	59.43$\pm$12.51 &	0.31$\pm$0.09	 &28.89$\pm$12.11	 &1.17$\pm$0.10	 &25.60$\pm$5.25	 &3.44$\pm$0.69 \\
\hline
F (1,8) &  2.06 &	0.42 &	4.69 &	0.35 &	0.18 &	5.48 &	0.10	 &1.10 &	11.8 &	4.32 &	0.70 &	0.02  \\
\hline
p       &  0.19 &	0.54 &	0.06 &	0.57 &	0.68 &	0.05 &	0.75 &	0.32 &	\textbf{\textless{}0.05}  &	0.07 &	0.43 &	0.90	\\
\hline \hline
\textbf{KT/Natural}  &   &  &  &    &    &  &    &     &  &     &    &  \\
\hline
Presence &  30.52$\pm$11.72	& 98.52$\pm$24.13	& 207.45$\pm$62.36	& 2.79$\pm$0.96& 	0.08$\pm$0.03	& 510.04$\pm$254.49& 	52.54$\pm$16.01& 	0.22$\pm$0.06& 	68.56$\pm$29.79& 	0.90$\pm$0.26& 	32.89$\pm$9.29& 	2.29$\pm$0.74	\\
\hline
Absence  & 23.02$\pm$9.5 &	78.75$\pm$28.25 &	110.45$\pm$40.67	 &1.83$\pm$1.0	 &0.03$\pm$0.01 &	100.37$\pm$81.72 &	33.49$\pm$17.02 &	0.09$\pm$0.04	 &9.56$\pm$7.1 &	0.57$\pm$0.28 &	18.96$\pm$5.9	 &1.08$\pm$0.52\\
\hline
F (1,8)   &  2.75 &	0.88 &	5.25 &	3.87 &	4.02 &	4.35 &	4.11 &	6.13 &	5.29 &	2.13 &	2.17 &	2.26\\
\hline
p         &   0.14 &	0.38 &	0.05 &	0.08 &	0.08 &	0.07	 &0.08	 &\textbf{\textless{}0.05} 	 &0.05 &	0.18 &	0.18 &	0.17\\                    

\hline \hline
\textbf{Video/Narration}  &   &  &  &    &    &  &    &     &  &     &    &  \\
\hline
Presence & 	90.41$\pm$2.55 &	149.27$\pm$12.35 &	302.99$\pm$46.52 &	4.96$\pm$0.79 &	0.12$\pm$0.03 &	497.48$\pm$217.64 &	96.71$\pm$8.58 &	0.33$\pm$0.04 &	89.06$\pm$34.77 &	2.15$\pm$0.14	 &9.66$\pm$2.52 &	3.52$\pm$0.23\\
\hline
Absence  & 	41.59$\pm$7.93 &	73.4$\pm$17.97 &	129.66$\pm$32.53 &	2.01$\pm$1.34 &	0.03$\pm$0.01 &	28.34$\pm$24.32 &	36.74$\pm$16.91 &	0.13$\pm$0.03 &	2.06$\pm$0.61	 &1.13$\pm$0.24 &	36.22$\pm$7.64 &	2.82$\pm$0.77 \\
\hline
F (1,8)   &   28.81 &	11.06 &	10.87 &	6.4	 &29.27	 &4.18 &	16.74 &	71.08 &	5.55 &	19.47 &	7.26 &	0.66       \\
\hline
p         & \textbf{\textless{}0.01}  &\textbf{\textless{}0.05} 	 &\textbf{\textless{}0.05}  &	\textbf{\textless{}0.05}  &	\textbf{\textless{}0.01}  &	0.08 &	\textbf{\textless{}0.01}  &	\textbf{\textless{}0.01}  &	0.05 &	\textbf{\textless{}0.01} &	\textbf{\textless{}0.05}  &	0.44 	\\                    

\hline \hline
\textbf{Video/Natural}  &   &  &  &    &    &  &    &     &  &     &    &  \\
\hline
Presence &  50.0$\pm$15.04 &	92.71$\pm$29.52 &	151.67$\pm$50.26 &	2.47$\pm$0.94 &	0.04$\pm$0.01 &	178.38$\pm$100.9 &	52.64$\pm$18.06 &	0.14$\pm$0.04 &	35.94$\pm$14.71 &	1.29$\pm$0.39 &	5.56$\pm$2.60 &	2.09$\pm$0.67	\\
\hline
Absence  & 21.29$\pm$9.09	 &78.09$\pm$31.71	 &109.48$\pm$44.84 &	2.61$\pm$1.21 &	0.02$\pm$0.01 &	44.8$\pm$36.69 &	44.41$\pm$19.69 &	0.10$\pm$0.04	 &1.94$\pm$1.05 &	0.61$\pm$0.33 &	26.0$\pm$9.20 &	1.54$\pm$0.79\\
\hline
F (1,8)   &    7.11 &	1.08 &	4.00 &	0.02 &	1.55 &	3.56 &	0.33 &	1.49 &	5.31 &	3.22 &	6.53 &	0.62     \\
\hline
p         & \textbf{\textless{}0.05}  & 0.33 &	0.08 &	0.88 &	0.25 &	0.10 &	0.58 &	0.26 &	0.05 &	0.11 &	\textbf{\textless{}0.05}&	0.45   \\                    

\hline
\end{tabular}}
\label{tab:stats_prosody_subtasks}
\end{table*}

\begin{table*}
    \centering
    \caption{Means and standard errors of speech density, hand-crafted acoustic features (Audio II), GloVe embeddings in the presence and absence of errors.
    }
    \resizebox{\textwidth}{!}{
    \begin{tabular}{|c||c||c|c|c|c|c|c|c|c|c|c||c|}
    \hline
 & \specialcell{Speech\\Density (\%)} & \specialcell{Mean\\Pitch} & \specialcell{Max\\Pitch} & \specialcell{Mean\\Energy(1E-4)}  & \specialcell{Max\\Energy} & \specialcell{Total\\Energy} & \specialcell{Mean\\Loudness(1E-4)} & \specialcell{Max\\Loudness} &  \specialcell{Word\\Count} & \specialcell{Word\\Rate} &
 \specialcell{Pause\\Density (\%)}&
 \specialcell{GloVe \\(PC Variance)}  \\
 \hline
\textbf{KT/Narration}  &   &  &  &    &    &  &    &     &  &     &    &  \\
\hline
Presence      & 57.33$\pm$7.58 &	135.44$\pm$12.44 &	284.46$\pm$36.31 &	2.58$\pm$0.62 &	0.25$\pm$0.11 &	217.14$\pm$76.44 &	45.12$\pm$6.57 &	0.41$\pm$0.09 &	36.89$\pm$13.57 &	0.94$\pm$0.13 &	38.26$\pm$5.78 &	4.44$\pm$0.7\\
\hline
Absence   & 70.12$\pm$5.32 &	139.82$\pm$10.66 &	317.04$\pm$44.02 &	3.28$\pm$0.65 &	0.17$\pm$0.04 &	1139.48$\pm$391.22 &	65.83$\pm$8.21 &	0.38$\pm$0.05 &	177.61$\pm$46.27 &	1.27$\pm$0.13 &	27.98$\pm$4.37 &	3.77$\pm$0.23\\
\hline
F (1,8) &    6.74 &	0.28 &	2.07 &	2.89 &	0.39 &	7.15 &	22.04 &	0.05 &	13.49 &	14.63 &	5.55 &	1.31 \\
\hline
p       & \textbf{\textless{}0.05}  &	0.61 &	0.19 &	0.13 &	0.55 &	\textbf{\textless{}0.05}  &	\textbf{\textless{}0.01}  &	0.82 &	\textbf{\textless{}0.05}  &	\textbf{\textless{}0.05}  &	0.05 &	0.29   \\
\hline \hline
\textbf{KT/Natural}  &   &  &  &    &    &  &    &     &  &     &    &  \\
\hline
Presence & 20.88$\pm$10.42 &	81.7$\pm$28.96 &	125.3$\pm$46.47 &	1.66$\pm$0.68	 &0.02$\pm$0.01 &	64.77$\pm$55.97 &	36.14$\pm$13.44 &	0.11$\pm$0.04 &	7.61$\pm$6.13	 &0.63$\pm$0.22 &	16.54$\pm$8.26 &	2.1$\pm$1.06\\
\hline
Absence  & 31.77$\pm$11.80 &	98.38$\pm$24.00 &	207.45$\pm$62.36 &	2.71$\pm$0.97	 &0.08$\pm$0.03 &	545.64$\pm$280.75	 &50.4$\pm$16.28 &	0.21$\pm$0.06	 &71.89$\pm$32.11	 &0.86$\pm$0.26 &	27.38$\pm$7.45 &	2.41$\pm$0.8\\
\hline
F (1,8)   &    2.64 &	1.99 &	5.95 &	1.74 &	5.11 &	3.71 &	1.14 &	5.64 &	4.80 &	1.58 &	0.96 &	0.15            \\
\hline
p         & 0.14 &	0.20 &	\textbf{\textless{}0.05}  &	0.22 &	0.05 &	0.09 &	0.32 &\textbf{\textless{}0.05}  &	0.06	 &0.24 &	0.36 &	0.71  \\           
\hline
\end{tabular}}
\label{tab:stats_prosody_errors}
\end{table*}

\subsection{Human Audio Analysis in relation with Subtasks and Errors during Demonstrations (\textbf{H3}, \textbf{H4})}
\label{sec:err_seg}
The percentage of demonstrations that consist of subtask segments and errors for both IVs are shown in the last 2 columns of Table~\ref{tab:stats}. The percentage for subtasks is significantly different across demonstration type (KT: $M=89.43, SD=8.81$; video: $M=83.09, SD=7.17$; $F(1,16)=6.78, p<0.05$) but not across instruction type (Narration: $M=87.06, SD=7.84$; Natural: $M=85.55, SD=9.36$; $F(1,16)=0.36, p=0.55$). The former effect can be explained by the presence of more errors, pauses, and gaps during KT demonstrations (Sec.~\ref{sec:audio_prosody}). The latter effect indicates that the execution of subtasks does not vary based on the instruction for using audio.  
The percentage of errors is very low for video demonstrations ($M=4.85, SD=6.10$), resulting in a negligible sample size of errors. Hence, we didn't analyze errors for video demonstrations any further.  
For KT, the percentage of errors (M=19.50, SD=9.23) are not significantly different across instruction types (Narration: $M=22.30, SD=16.98$; Natural: $M=16.70, SD=6.16$; $F(1,16)=0.6, p=0.45$). For further analysis, we perform 1-way ANOVA analyses using four subsets of our data independently. 
The four subsets are the conditions that result from combinations of demonstration type and instruction type. For each subset, a single IV is a binary category depicting the presence or absence of subtasks/errors. The DMs are same as before (Sec.~\ref{sec:dv}).

\subsubsection{Quantification of 
Utterances}
\label{sec:seg_err_dur}
As shown in Table~\ref{tab:stats_prosody_errors}, for each subset of the data, the 
density of speech is higher during absence of errors, significantly more ($F(1,8)=6.74, p<0.05$) during kinesthetic demonstrations versus video demonstrations under the narration condition. This finding does not provide support for H4.
Talking more during absence of errors can be explained by the fact that majority of the demonstration data does not comprise of errors (Table~\ref{tab:stats}). However, there are still interesting acoustic cues present \textit{during} errors as shown in Table~\ref{tab:stats_annot_errors}.

A similar trend of 
higher speech density is observed \textit{during} subtasks in comparison to in their absence, with significant results under subsets of video demonstrations (Video/Narration: $F(1,8)=28.81, p<0.01$; Video/Natural: $F(1,8)=7.11, p<0.05$) 
as shown in Table~\ref{tab:stats_prosody_subtasks}. For kinesthetic demonstrations, users talk roughly as much in the presence of subtasks as in their absence.
These findings provide partial support for H3. 

\subsubsection{Speech Prosody}
\label{sec:prosody_err_seg}

For analysis of subtasks, we observe that most acoustic features are higher \textit{during} subtask presence versus absence (Table~\ref{tab:stats_prosody_subtasks}), with most being statistically significant for video demonstrations under the narration condition. 
However, as shown in Table~\ref{tab:stats_prosody_subtasks}, teachers still exhibit more expressive speech during gaps between subtask executions (such as surprise and laughter during KT demonstrations in the narration condition). This result is similar to Nagai et al. \cite{nagai2008toward}, where demonstrators use expressive infant-directed speech before they begin the execution of a subtask.
As a few of the results in Table~\ref{tab:stats_prosody_subtasks} and Table~\ref{tab:stats_annot_subtasks} are statistically significant (more so during the narration condition), they only provide partial support for H3. 

For analysis of errors (Table~\ref{tab:stats_prosody_errors}), we observe that most prosodic features are higher in the absence of errors, since the majority of demonstrations comprise of non-erroneous segments (Table~\ref{tab:stats}). 
However, there are still interesting acoustic cues such as frustration, surprise, pauses, laughter, and encouragement present \textit{during} errors as shown in Table~\ref{tab:stats_annot_errors}. This
indicates that teachers convey some of their reactions with emphasis when a mistake happens. This finding provides only partial support for H4.

\begin{table}
    \centering
    \caption{Means and standard errors of annotated acoustic feature density (\%) (Audio I) in the presence and absence of subtasks. 
    }
    \resizebox{0.48\textwidth}{!}{
    \begin{tabular}{|c||c|c|c|c|c|c|}
    \hline
 & \specialcell{Frustration} & 
 \specialcell{Surprise}&
 \specialcell{Speech\\Pauses} &   \specialcell{Normal\\Speech} &
 \specialcell{Laughter}  & \specialcell{Encourage\\-ment}\\
\hline
\textbf{KT/Narration}        &  &       &     &   & &    \\
\hline
Presence       &  4.76$\pm$1.88 &	6.71$\pm$2.33 &	4.02$\pm$1.93 &	90.89$\pm$4.12 &	1.86$\pm$1.75 &	6.27$\pm$3.01\\
\hline
Absence    &   0.00$\pm$0.00	 &10.49$\pm$7.76 &	0.3$\pm$0.29 &	14.17$\pm$10.07 &	2.69$\pm$2.53 &	0.03$\pm$0.03\\
\hline
F(1,8)   &  5.70 &	0.24 &	3.11 &	50.09 &	1.00 &	3.83   \\
\hline
p         &  \textbf{\textless{}0.05}	 &0.64	 &0.12 &	\textbf{\textless{}0.01} &	0.35 &	0.09 \\
\hline \hline
\textbf{KT/Natural}       &  &       &     &   & &    \\
\hline
Presence   &  0.00$\pm$0.00 &	0.86$\pm$0.69 &	5.64$\pm$2.31 &	46.72$\pm$13.79 &	0.00$\pm$0.00 &	0.3$\pm$0.28  \\
\hline
Absence & 0.00$\pm$0.00 &	0.00$\pm$0.00 &	1.72$\pm$1.62&	5.62$\pm$5.20 &	0.00$\pm$0.00 &	0.00$\pm$0.00 \\
\hline
F(1,8)   &   NA &	1.38 &	2.41 &	9.96 &	NA &	1.00 \\
\hline
p         & NA &	0.27 &	0.16 &	\textbf{\textless{}0.05} &	NA &	0.35\\
\hline \hline
\textbf{Video/Narration}       &  &       &     &   & &    \\
\hline
Presence   &  0.00$\pm$0.00 &	0.20$\pm$0.19 &	0.00$\pm$0.00 &	91.71$\pm$3.73 &	0.00$\pm$0.00 &	0.00$\pm$0.00  \\
\hline
Absence &   0.00$\pm$0.00 &	0.00$\pm$0.00 &	0.00$\pm$0.00 &	0.48$\pm$0.46 &	0.00$\pm$0.00 &	0.00$\pm$0.00 \\
\hline
F(1,8)   &   NA &	1.00 &	NA &	556.53 &	NA &	NA \\
\hline
p         & NA &	0.35 &	NA &	\textbf{\textless{}0.01} &	NA &	NA\\
\hline \hline
\textbf{Video/Natural}       &  &       &     &   & &    \\
\hline
Presence   &  0.00$\pm$0.00 &	0.00$\pm$0.00 &	0.00$\pm$0.00 &	52.12$\pm$15.59 &	0.00$\pm$0.00	 &0.00$\pm$0.00 \\
\hline
Absence &   0.00$\pm$0.00	 &0.00$\pm$0.00 &	0.00$\pm$0.00 &	0.00$\pm$0.00	 &0.00$\pm$0.00 &	0.00$\pm$0.00  \\
\hline
F(1,8)   &   NA &	NA &	NA &	9.94 &	NA &	NA\\
\hline
p         & NA	&NA &	NA &	\textbf{\textless{}0.05} &	NA &	NA\\
\hline
\end{tabular}}
\label{tab:stats_annot_subtasks}
\end{table}

\begin{table}
    \centering
    \caption{Means and standard errors of annotated acoustic feature density (\%) (Audio I) in the presence and absence of errors. 
    }
    \resizebox{0.48\textwidth}{!}{
    \begin{tabular}{|c||c|c|c|c|c|c|}
    \hline
 & \specialcell{Frustration} & 
 \specialcell{Surprise}&
 \specialcell{Speech\\Pauses} &   \specialcell{Normal\\Speech} &
 \specialcell{Laughter}  & \specialcell{Encourage\\-ment}\\
\hline
\textbf{KT/Narration}        &  &       &     &   & &    \\
\hline
Presence        &15.1$\pm$7.18 &	19.04$\pm$9.29 &	15.51$\pm$6.6 &	84.68$\pm$5.2	 &5.69$\pm$5.37 &	10.23$\pm$6.71 \\
\hline
Absence    & 1.15$\pm$0.44 &	3.68$\pm$1.93 &	0.0$\pm$0.0 &	78.4$\pm$6.03 &	0.23$\pm$0.22 &	4.12$\pm$2.56  \\
\hline
F(1,8)   &  3.62 &	2.34 &	4.91 &	0.70 &	1.00 &	0.81  \\
\hline
p         &  0.09 &	0.16 &	0.06 &	0.43 &	0.35 &	0.39\\
\hline \hline
\textbf{KT/Natural}        &  &       &     &   & &    \\
\hline
Presence   &   0.00$\pm$0.00 &	1.25$\pm$1.18 &	9.86$\pm$8.78 &	35.05$\pm$13.96 &	0.00$\pm$0.00 &	0.27$\pm$0.26  \\
\hline
Absence &   0.00$\pm$0.00 &	1.07$\pm$0.99 &	0.01$\pm$0.01 &	37.9$\pm$12.83 &	0.00$\pm$0.00 &	0.19$\pm$0.18\\
\hline
F(1,8)   &   NA &	0.01 &	1.12 &	0.21 &	NA &	1.00 \\
\hline
p         & NA &	0.92 &	0.32 &	0.66 &	NA &	0.35\\
\hline
\end{tabular}}
\label{tab:stats_annot_errors}
\end{table}

\subsubsection{Information Conveyed via Spoken Words}
We do not observe a significant difference in the variance of GloVe features under any of the data subsets for subtask analysis (KT/Narration: $F(1,8)=0.02,p=0.90$; KT/Natural: $F(1,8)=2.26,p=0.17$; Video/Narration: $F(1,8)=0.66,p=0.44$; Video/Natural: $F(1,8)=0.62,p=0.45$) or error analysis (KT/Narration: $F(1,8)=1.31,p=0.29$, KT/Natural $F(1,8)=0.15,p=0.71$). 
This implies that the concepts being communicated in the presence or absence of subtasks and errors are quite similar. However, a few words are specifically spoken under certain conditions (such as the use of `oops$^^21$', `oh oh' etc. when errors occur). 

\subsection{Learning Experiments}
\label{sec:learning-results} 

\begin{table}
    \centering
    \footnotesize
    \caption{Average F1 scores of Subtask Detection and Error Detection for users in the `Narration' instruction condition. 
    }
    \resizebox{0.47\textwidth}{!}{
    \begin{tabular}{|c||c|c|c|c||c|c|}
    \hline
    & \multicolumn{4}{c||}{Subtask Detection} & \multicolumn{2}{c|}{Error Detection} \\
    \hline
    Features & \begin{tabular}[c]{@{}c@{}}Box\\ Video\end{tabular} & \begin{tabular}[c]{@{}c@{}}Box\\ KT\end{tabular}  &
    \begin{tabular}[c]{@{}c@{}}Cutting\\ Video\end{tabular} & \begin{tabular}[c]{@{}c@{}}Cutting\\ KT\end{tabular} & \begin{tabular}[c]{@{}c@{}}Box\\ KT\end{tabular}  & \begin{tabular}[c]{@{}c@{}}Cutting\\ KT\end{tabular} \\
    \hline
        \hline
        Random             & 0.44                                   & 0.4                                 & 0.44                                       & 0.43                                    & 0.43                                & 0.47                                    \\
Constant           & 0.46                                   & 0.47                                & 0.46                                       & 0.46                                    & 0.46                                & 0.42                                    \\
\hline
\hline
Audio I            & 0.86                                   & 0.84                                & 0.76                                       & 0.78                                    & 0.89                                & 0.80                                     \\
Audio II           & 0.71                                   & 0.60                                 & 0.72                                       & 0.70                                     & 0.56                                & 0.55                                    \\
Audio (I+II)       & \textbf{0.91}                          & \textbf{0.87}                       & \textbf{0.79}                              & \textbf{0.89}                           & \textbf{0.89}                       & \textbf{0.82}                           \\
PASE               & 0.59                                   & 0.49                                & 0.71                                       & 0.50                                     & 0.46                                & 0.46                                    \\
\hline
\hline
Video              & 0.82                                   & 0.80                                 & 0.66                                       & 0.74                                    & 0.72                                & 0.67                                    \\
Audio I+Video      & 0.78                                   & 0.84                                & 0.77                                       & 0.80                                     & \textbf{0.90}                        & \textbf{0.84}                           \\
Audio II+Video     & 0.80                                    & 0.82                                & 0.53                                       & 0.77                                    & 0.69                                & 0.66                                    \\
Audio (I+II)+Video & 0.82                                   & 0.83                                & 0.76                                       & 0.78                                    & \textbf{0.90}                        & \textbf{0.84}     \\                                      
    \hline
    \end{tabular}}
    \label{tab:results_seg_presence}
\end{table}

From results of Sec.~\ref{sec:seg_err_dur} and Sec.~\ref{sec:prosody_err_seg}, we observe that the narration condition has more significant results for acoustic feature differences in relation to subtasks and errors
. Thus, we use acoustic features from the narration condition to evaluate performance on two binary classification tasks: subtask detection and error detection.
Random $80\%-20\%$ splits are used to respectively train and test random forest classifiers. 
We use a combination of acoustic and video features as input to the model.
Two baseline models (random prediction and best constant class prediction) are also evaluated.
We train three random forests per experiment, and report the average F1 score on the test data for best of three runs in 
Table~\ref{tab:results_seg_presence}. 
Our results show that acoustic features perform better than random and constant class prediction baselines. Combining both annotated (Audio I) and hand-crafted (Audio II) acoustic features is better than using either one alone (for both subtask detection and error detection). Pretrained PASE features (no finetuning) perform only slightly better than the baselines, and are unable to match the performance of Audio I and Audio II features. 
In addition, pretrained video features (no finetuning) are also not as effective alone as the acoustic features for either detection task. 
However, combining acoustic and video features performs better at error detection than using acoustic or video features alone.
For subtask detection, we find that concatenating the annotated and hand-crafted acoustic features gives the best performance.
These results highlight that acoustic features contain rich information about demonstration quality and task segments, with the potential to enhance the performance of other LfD  approaches.
\section{Conclusion}

Our work highlights several characteristics of audio signals exhibited by human teachers providing demonstrations for multi-step manipulation tasks to a situated robot. Our findings indicate that human audio cues carry rich information, potentially beneficial for further technological advancement in robot learning.
While several human demonstration datasets leverage environment and object sounds~\cite{zhang2019cutting, gandhi2019swoosh,liang2019making,schaal2006dynamic},
often human audio of demonstrators is not recorded. Since speech data can be recorded easily with light-weight and cheap sensors, 
we propose that collection of human audio data, as part of future demonstration datasets, can be beneficial for algorithm development. Integrating information from (1) environmental and object sounds along with (2) information from human audio is a topic for future work.  Our analysis is done on a small in-house dataset collected in a laboratory setting. Leveraging human audio for larger datasets with larger, data-driven feature encoders, such as neural networks, is also a topic for future work.
Furthermore, several recent LfD approaches use suboptimal demonstrations~\cite{brown2019extrapolating,gao2018reinforcement} as input. Real-world demonstration data can contain partial errors and gaps in between execution steps of a task. Instead of discarding suboptimal data, leveraging an additional modality like audio can provide additional information about errors and subtasks to aid learning.
We take the key first step in leveraging human audio for robot learning---understanding the information present in speech and highlighting that it is possible for automated methods to extract it.

\section*{ACKNOWLEDGMENTS}
The authors would like to thank Sahil Jain and Dr. Greg Trafton for help with data analysis, Prof. Elaine Schaertl Short for providing valuable feedback on the draft, Prof. Sonia Chernova and Prof. Maya Cakmak for helpful discussions about prior work. This work has partly taken place in the Personal Autonomous Robotics Lab (PeARL) at The University of Texas at Austin. PeARL research is supported in part by the NSF (IIS-1724157, IIS-1749204, IIS-1925082), AFOSR (FA9550-20-1-0077), ARO (78372-CS), DFG (448648559).  This research was also sponsored by the Army Research Office under Cooperative Agreement Number W911NF-19-2-0333. The views and conclusions contained in this document are those of the authors and should not be interpreted as representing the official policies, either expressed or implied, of the Army Research Office or the U.S. Government. The U.S. Government is authorized to reproduce and distribute reprints for Government purposes notwithstanding any copyright notation herein.

\footnotesize
\bibliographystyle{IEEEtran}
\bibliography{root}

\begin{thebibliography}{10}
\providecommand{\url}[1]{#1}
\csname url@rmstyle\endcsname
\providecommand{\newblock}{\relax}
\providecommand{\bibinfo}[2]{#2}
\providecommand\BIBentrySTDinterwordspacing{\spaceskip=0pt\relax}
\providecommand\BIBentryALTinterwordstretchfactor{4}
\providecommand\BIBentryALTinterwordspacing{\spaceskip=\fontdimen2\font plus
\BIBentryALTinterwordstretchfactor\fontdimen3\font minus
  \fontdimen4\font\relax}
\providecommand\BIBforeignlanguage[2]{{%
\expandafter\ifx\csname l@#1\endcsname\relax
\typeout{** WARNING: IEEEtran.bst: No hyphenation pattern has been}%
\typeout{** loaded for the language `#1'. Using the pattern for}%
\typeout{** the default language instead.}%
\else
\language=\csname l@#1\endcsname
\fi
#2}}

\bibitem{oviatt2004we}
S.~Oviatt, R.~Coulston, and R.~Lunsford, ``When do we interact multimodally?
  {C}ognitive load and multimodal communication patterns,'' in
  \emph{Proceedings of the 6th International Conference on Multimodal
  Interfaces (ICMI)}, 2004, pp. 129--136.

\bibitem{argall2009survey}
B.~D. Argall, S.~Chernova, M.~Veloso, and B.~Browning, ``A survey of robot
  learning from demonstration,'' \emph{Robotics and Autonomous Systems},
  vol.~57, no.~5, pp. 469--483, 2009.

\bibitem{osa2018algorithmic}
T.~Osa, J.~Pajarinen, G.~Neumann, J.~A. Bagnell, P.~Abbeel, and J.~Peters, ``An
  algorithmic perspective on imitation learning,'' \emph{arXiv preprint
  arXiv:1811.06711}, 2018.

\bibitem{saran2018human}
A.~Saran, S.~Majumdar, E.~S. Short, A.~Thomaz, and S.~Niekum, ``Human gaze
  following for human-robot interaction,'' in \emph{2018 IEEE/RSJ International
  Conference on Intelligent Robots and Systems (IROS)}.\hskip 1em plus 0.5em
  minus 0.4em\relax IEEE, 2018, pp. 8615--8621.

\bibitem{saran2020understanding}
A.~Saran, E.~S. Short, A.~Thomaz, and S.~Niekum, ``Understanding teacher gaze
  patterns for robot learning,'' in \emph{Conference on Robot Learning}.\hskip
  1em plus 0.5em minus 0.4em\relax PMLR, 2020, pp. 1247--1258.

\bibitem{zhang2020human}
R.~Zhang, A.~Saran, B.~Liu, Y.~Zhu, S.~Guo, S.~Niekum, D.~Ballard, and
  M.~Hayhoe, ``Human gaze assisted artificial intelligence: A review,'' in
  \emph{IJCAI: Proceedings of the Conference}, vol. 2020.\hskip 1em plus 0.5em
  minus 0.4em\relax NIH Public Access, 2020, p. 4951.

\bibitem{saran2020efficiently}
A.~Saran, R.~Zhang, E.~S. Short, and S.~Niekum, ``Efficiently guiding imitation
  learning agents with human gaze,'' \emph{International Conference on
  Autonomous Agents and Multiagent Systems}, 2021.

\bibitem{yu2016automatic}
D.~Yu and L.~Deng, \emph{AUTOMATIC SPEECH RECOGNITION.}\hskip 1em plus 0.5em
  minus 0.4em\relax Springer, 2016.

\bibitem{clark2002using}
H.~H. Clark and J.~E.~F. Tree, ``Using uh and um in spontaneous speaking,''
  \emph{Cognition}, vol.~84, no.~1, pp. 73--111, 2002.

\bibitem{scassellati2009affective}
B.~Scassellati, ``Affective prosody recognition for human-robot interaction,''
  in \emph{Microsoft Research’s External Research Symposium. Redmond, WA,
  USA}.\hskip 1em plus 0.5em minus 0.4em\relax Citeseer, 2009.

\bibitem{kim2009people}
E.~S. Kim, D.~Leyzberg, K.~M. Tsui, and B.~Scassellati, ``How people talk when
  teaching a robot,'' in \emph{Proceedings of the 4th ACM/IEEE International
  Conference on Human-Robot Interaction (HRI)}, 2009, pp. 23--30.

\bibitem{short2018detecting}
E.~S. Short, M.~L. Chang, and A.~Thomaz, ``Detecting contingency for hri in
  open-world environments,'' in \emph{Proceedings of the 2018 ACM/IEEE
  International Conference on Human-Robot Interaction}, 2018, pp. 425--433.

\bibitem{nagai2008toward}
Y.~Nagai, C.~Muhl, and K.~J. Rohlfing, ``Toward designing a robot that learns
  actions from parental demonstrations,'' in \emph{2008 IEEE international
  conference on robotics and automation}.\hskip 1em plus 0.5em minus
  0.4em\relax IEEE, 2008, pp. 3545--3550.

\bibitem{nakamura2015constructing}
R.~Nakamura, K.~Miyazawa, H.~Ishihara, K.~Nishikawa, H.~Kikuchi, M.~Asada, and
  R.~Mazuka, ``Constructing the corpus of infant-directed speech and
  infant-like robot-directed speech,'' in \emph{Proceedings of the 3rd
  International Conference on Human-Agent Interaction}, 2015, pp. 167--169.

\bibitem{nicolescu2003natural}
M.~N. Nicolescu and M.~J. Mataric, ``Natural methods for robot task learning:
  Instructive demonstrations, generalization and practice,'' in
  \emph{Proceedings of the second International Joint Conference on Autonomous
  Agents and Multiagent Systems}, 2003, pp. 241--248.

\bibitem{pardowitz2007incremental}
M.~Pardowitz, S.~Knoop, R.~Dillmann, and R.~D. Zollner, ``Incremental learning
  of tasks from user demonstrations, past experiences, and vocal comments,''
  \emph{IEEE Transactions on Systems, Man, and Cybernetics, Part B
  (Cybernetics)}, vol.~37, no.~2, pp. 322--332, 2007.

\bibitem{tenorio2010dynamic}
A.~C. Tenorio-Gonzalez, E.~F. Morales, and L.~Villase{\~n}or-Pineda, ``Dynamic
  reward shaping: training a robot by voice,'' in \emph{Ibero-American
  Conference on Artificial Intelligence}.\hskip 1em plus 0.5em minus
  0.4em\relax Springer, 2010, pp. 483--492.

\bibitem{kim2007learning}
E.~S. Kim and B.~Scassellati, ``Learning to refine behavior using prosodic
  feedback,'' in \emph{2007 IEEE 6th International Conference on Development
  and Learning}.\hskip 1em plus 0.5em minus 0.4em\relax IEEE, 2007, pp.
  205--210.

\bibitem{krening2016learning}
S.~Krening, B.~Harrison, K.~M. Feigh, C.~L. Isbell, M.~Riedl, and A.~Thomaz,
  ``Learning from explanations using sentiment and advice in rl,'' \emph{IEEE
  Transactions on Cognitive and Developmental Systems}, vol.~9, no.~1, pp.
  44--55, 2016.

\bibitem{krening2018newtonian}
S.~Krening, ``Newtonian action advice: Integrating human verbal instruction
  with reinforcement learning,'' \emph{arXiv preprint arXiv:1804.05821}, 2018.

\bibitem{kroemer2019review}
O.~Kroemer, S.~Niekum, and G.~Konidaris, ``A review of robot learning for
  manipulation: Challenges, representations, and algorithms,'' \emph{arXiv
  preprint arXiv:1907.03146}, 2019.

\bibitem{quigley2009ros}
M.~Quigley, K.~Conley, B.~Gerkey, J.~Faust, T.~Foote, J.~Leibs, R.~Wheeler,
  A.~Y. Ng, \emph{et~al.}, ``Ros: an open-source robot operating system,'' in
  \emph{ICRA workshop on open source software}, vol.~3, no. 3.2.\hskip 1em plus
  0.5em minus 0.4em\relax Kobe, Japan, 2009, p.~5.

\bibitem{akgun2012trajectories}
B.~Akgun, M.~Cakmak, J.~W. Yoo, and A.~L. Thomaz, ``Trajectories and keyframes
  for kinesthetic teaching: A human-robot interaction perspective,'' in
  \emph{Proceedings of the seventh annual ACM/IEEE international conference on
  Human-Robot Interaction}, 2012, pp. 391--398.

\bibitem{fischer2016comparison}
K.~Fischer, F.~Kirstein, L.~C. Jensen, N.~Kr{\"u}ger, K.~Kukli{\'n}ski, M.~V.
  aus~der Wieschen, and T.~R. Savarimuthu, ``A comparison of types of robot
  control for programming by demonstration,'' in \emph{2016 11th ACM/IEEE
  International Conference on Human-Robot Interaction (HRI)}.\hskip 1em plus
  0.5em minus 0.4em\relax IEEE, 2016, pp. 213--220.

\bibitem{schillingmann2009computational}
L.~Schillingmann, B.~Wrede, and K.~J. Rohlfing, ``A computational model of
  acoustic packaging,'' \emph{IEEE Transactions on Autonomous Mental
  Development}, vol.~1, no.~4, pp. 226--237, 2009.

\bibitem{hirschberg2004prosodic}
J.~Hirschberg, D.~Litman, and M.~Swerts, ``Prosodic and other cues to speech
  recognition failures,'' \emph{Speech communication}, vol.~43, no. 1-2, pp.
  155--175, 2004.

\bibitem{juang2005automatic}
B.-H. Juang and L.~R. Rabiner, ``Automatic speech recognition--a brief history
  of the technology development,'' \emph{Georgia Institute of Technology.
  Atlanta Rutgers University and the University of California. Santa Barbara},
  vol.~1, p.~67, 2005.

\bibitem{benzeghiba2007automatic}
M.~Benzeghiba, R.~De~Mori, O.~Deroo, S.~Dupont, T.~Erbes, D.~Jouvet,
  L.~Fissore, P.~Laface, A.~Mertins, C.~Ris, \emph{et~al.}, ``Automatic speech
  recognition and speech variability: A review,'' \emph{Speech communication},
  vol.~49, no. 10-11, pp. 763--786, 2007.

\bibitem{googlestt}
Google, ``{Google Cloud Speech-to-Tex},''
  \url{https://cloud.google.com/speech-to-text}, 2021.

\bibitem{charmaz2007grounded}
K.~Charmaz and L.~L. Belgrave, ``Grounded theory,'' \emph{The Blackwell
  encyclopedia of sociology}, 2007.

\bibitem{charmaz2014constructing}
K.~Charmaz, \emph{Constructing grounded theory}.\hskip 1em plus 0.5em minus
  0.4em\relax sage, 2014.

\bibitem{morse2016developing}
J.~M. Morse, B.~J. Bowers, K.~Charmaz, J.~Corbin, A.~E. Clarke, and P.~N.
  Stern, \emph{Developing grounded theory: The second generation}.\hskip 1em
  plus 0.5em minus 0.4em\relax Routledge, 2016, vol.~3.

\bibitem{defossez2020real}
A.~Defossez, G.~Synnaeve, and Y.~Adi, ``Real time speech enhancement in the
  waveform domain,'' \emph{arXiv preprint arXiv:2006.12847}, 2020.

\bibitem{tran2017parsing}
T.~Tran, S.~Toshniwal, M.~Bansal, K.~Gimpel, K.~Livescu, and M.~Ostendorf,
  ``Parsing speech: a neural approach to integrating lexical and
  acoustic-prosodic information,'' \emph{arXiv preprint arXiv:1704.07287},
  2017.

\bibitem{kim2018crepe}
J.~W. Kim, J.~Salamon, P.~Li, and J.~P. Bello, ``Crepe: A convolutional
  representation for pitch estimation,'' in \emph{2018 IEEE International
  Conference on Acoustics, Speech and Signal Processing (ICASSP)}.\hskip 1em
  plus 0.5em minus 0.4em\relax IEEE, 2018, pp. 161--165.

\bibitem{pascual2019learning}
S.~Pascual, M.~Ravanelli, J.~Serra, A.~Bonafonte, and Y.~Bengio, ``Learning
  problem-agnostic speech representations from multiple self-supervised
  tasks,'' \emph{arXiv preprint arXiv:1904.03416}, 2019.

\bibitem{pennington2014glove}
J.~Pennington, R.~Socher, and C.~D. Manning, ``Glove: Global vectors for word
  representation,'' in \emph{Proceedings of the 2014 conference on empirical
  methods in natural language processing (EMNLP)}, 2014, pp. 1532--1543.

\bibitem{pedregosa2011scikit}
F.~Pedregosa, G.~Varoquaux, A.~Gramfort, V.~Michel, B.~Thirion, O.~Grisel,
  M.~Blondel, P.~Prettenhofer, R.~Weiss, V.~Dubourg, \emph{et~al.},
  ``Scikit-learn: Machine learning in python,'' \emph{the Journal of machine
  Learning research}, vol.~12, pp. 2825--2830, 2011.

\bibitem{carreira2017quo}
J.~Carreira and A.~Zisserman, ``Quo vadis, action recognition? a new model and
  the kinetics dataset,'' in \emph{proceedings of the IEEE Conference on
  Computer Vision and Pattern Recognition}, 2017, pp. 6299--6308.

\bibitem{kay2017kinetics}
W.~Kay, J.~Carreira, K.~Simonyan, B.~Zhang, C.~Hillier, S.~Vijayanarasimhan,
  F.~Viola, T.~Green, T.~Back, P.~Natsev, \emph{et~al.}, ``The kinetics human
  action video dataset,'' \emph{arXiv preprint arXiv:1705.06950}, 2017.

\bibitem{zhang2019cutting}
K.~Zhang, M.~Sharma, M.~Veloso, and O.~Kroemer, ``Leveraging multimodal haptic
  sensory data for robust cutting,'' in \emph{2019 IEEE-RAS 19th International
  Conference on Humanoid Robots (Humanoids)}.\hskip 1em plus 0.5em minus
  0.4em\relax IEEE, 2019, pp. 409--416.

\bibitem{gandhi2019swoosh}
D.~Gandhi, A.~Gupta, and L.~Pinto, ``Swoosh! rattle! thump!-actions that
  sound,'' in \emph{Robotics: Science and Systems (RSS)}, 2019.

\bibitem{liang2019making}
H.~Liang, S.~Li, X.~Ma, N.~Hendrich, T.~Gerkmann, F.~Sun, and J.~Zhang,
  ``Making sense of audio vibration for liquid height estimation in robotic
  pouring,'' \emph{arXiv preprint arXiv:1903.00650}, 2019.

\bibitem{schaal2006dynamic}
S.~Schaal, ``Dynamic movement primitives-a framework for motor control in
  humans and humanoid robotics,'' in \emph{Adaptive motion of animals and
  machines}.\hskip 1em plus 0.5em minus 0.4em\relax Springer, 2006, pp.
  261--280.

\bibitem{brown2019extrapolating}
D.~Brown, W.~Goo, P.~Nagarajan, and S.~Niekum, ``Extrapolating beyond
  suboptimal demonstrations via inverse reinforcement learning from
  observations,'' in \emph{International Conference on Machine Learning}.\hskip
  1em plus 0.5em minus 0.4em\relax PMLR, 2019, pp. 783--792.

\bibitem{gao2018reinforcement}
Y.~Gao, H.~Xu, J.~Lin, F.~Yu, S.~Levine, and T.~Darrell, ``Reinforcement
  learning from imperfect demonstrations,'' \emph{arXiv preprint
  arXiv:1802.05313}, 2018.

\end{thebibliography}

\end{document}